\documentclass{article}

\usepackage{microtype}
\usepackage{graphicx}
\usepackage{subfigure}
\usepackage{booktabs} %

\usepackage{caption}
\usepackage{color}
\usepackage{xspace}
\usepackage{amsfonts}       %
\usepackage{nicefrac}       %
\usepackage{color}
\usepackage{multirow}
\usepackage{multicol}
\usepackage{array}
\usepackage{amssymb,amsmath,bm,amsthm}
\usepackage[usenames,dvipsnames]{xcolor}
\usepackage{makecell}
\usepackage{enumitem}  %

\usepackage{amsmath,amsfonts,bm}

\def\eqref#1{equation~\ref{#1}}

\def\1{\bm{1}}

\DeclareMathAlphabet{\mathsfit}{\encodingdefault}{\sfdefault}{m}{sl}
\SetMathAlphabet{\mathsfit}{bold}{\encodingdefault}{\sfdefault}{bx}{n}

\def\gB{{\mathcal{B}}}

\def\gD{{\mathcal{D}}}

\def\gF{{\mathcal{F}}}

\def\gL{{\mathcal{L}}}

\def\gO{{\mathcal{O}}}

\usepackage{tabularx}
\newcolumntype{L}[1]{>{\raggedright\let\newline\\\arraybackslash\hspace{0pt}}m{#1}}
\newcolumntype{C}[1]{>{\centering\let\newline\\\arraybackslash\hspace{0pt}}m{#1}}
\newcolumntype{R}[1]{>{\raggedleft\let\newline\\\arraybackslash\hspace{0pt}}m{#1}}
\newcolumntype{Y}{>{\centering\arraybackslash}X}

\usepackage{pgfplots}
\DeclareUnicodeCharacter{2212}{−}
\usepgfplotslibrary{groupplots,dateplot}
\usetikzlibrary{patterns,shapes.arrows}
\pgfplotsset{compat=newest}

\definecolor{snsgray}{RGB}{179, 179, 179}
\definecolor{snsorange}{RGB}{252, 141, 98}
\definecolor{snsblue}{RGB}{141, 160, 203}

\definecolor{coolgrey}{RGB}{157,157,157}
\definecolor{lightgrey}{RGB}{235,238,238}
\definecolor{lightteal}{RGB}{198,211,222}
\definecolor{cyan}{RGB}{136, 204, 238}
\definecolor{teal}{RGB}{68, 170, 153}
\definecolor{sand}{RGB}{221, 204, 119}
\definecolor{rose}{RGB}{204, 102, 119}
\definecolor{red}{RGB}{250, 94, 91}
\definecolor{orange}{RGB}{255, 200, 63}
\definecolor{yellow}{RGB}{254, 239, 109}

\definecolor{darkgreen}{rgb}{0.09, 0.45, 0.27}

\newcommand{\narrowcolorbox}[2]{\setlength{\fboxsep}{2pt}\colorbox{#1}{#2}}

\newcommand{\secant}[0]{\mbox{\textsc{Secant}}\xspace}
\newcommand{\secantcaption}[0]{\colorbox{yellow}{\secant}}
\newcommand{\para}[1]{\noindent{\textbf{#1}}}
\newcommand{\bestscore}[1]{\textcolor{darkgreen}{\mathbf{#1}}}
\newcommand{\bestpercent}[1]{\textcolor{darkgreen}{\text{#1\%}}}
\newcommand{\specialcell}[2][c]{%
  \begin{tabular}[#1]{@{}c@{}}#2\end{tabular}}

\usepackage{hyperref}

\usepackage[accepted]{icml2021}

\newcommand{\MYTITLE}[0]{\secant: Self-Expert Cloning
           for Zero-Shot Generalization of Visual Policies}
\icmltitlerunning{\MYTITLE}

\begin{document}

\twocolumn[
\icmltitle{\MYTITLE}

\icmlsetsymbol{equal}{*}
\icmlsetsymbol{intern}{*}

\begin{icmlauthorlist}
\icmlauthor{Linxi Fan}{stan,nv,intern}
\icmlauthor{Guanzhi Wang}{stan}
\icmlauthor{De-An Huang}{nv}
\icmlauthor{Zhiding Yu}{nv}\\
\icmlauthor{Li Fei-Fei}{stan}
\icmlauthor{Yuke Zhu}{ut,nv}
\icmlauthor{Anima Anandkumar}{cal,nv}
\end{icmlauthorlist}

\icmlaffiliation{stan}{Stanford University, CA, USA.}
\icmlaffiliation{nv}{NVIDIA, CA, USA.}
\icmlaffiliation{ut}{The University of Texas at Austin, TX, USA.}
\icmlaffiliation{cal}{California Institute of Technology, CA, USA}

\icmlcorrespondingauthor{Linxi Fan}{jimfan@cs.stanford.edu}

\icmlkeywords{Reinforcement Learning, Computer Vision, Robust Learning, Sim-to-real, Robotics, ICML}

\vskip 0.3in
]

\printAffiliationsAndNotice{\icmlIntern}

\begin{abstract}

Generalization has been a long-standing challenge for reinforcement learning (RL). Visual RL, in particular, can be easily distracted by irrelevant factors in high-dimensional observation space. 
In this work, we consider robust policy learning which targets zero-shot generalization to unseen visual environments with large distributional shift.
We propose \secant, a novel self-expert cloning technique that leverages image augmentation in two stages to \textit{decouple} robust representation learning from policy optimization.
Specifically, an expert policy is first trained by RL from scratch with \textit{weak} augmentations. A student network then learns to mimic the expert policy by supervised learning with \textit{strong} augmentations, making its representation more robust against visual variations compared to the expert.
Extensive experiments demonstrate that \secant significantly advances the state of the art in zero-shot generalization across 4 challenging domains. Our average reward improvements over prior SOTAs are: DeepMind Control (+26.5\%), robotic manipulation (+337.8\%), vision-based autonomous driving (+47.7\%), and indoor object navigation (+15.8\%). Code release and video are available at \colorbox{yellow}{\href{https://linxifan.github.io/secant-site/}{this link}}.

\end{abstract}

\begin{figure}[t!]
    \centering
    \includegraphics[width=0.85\linewidth]{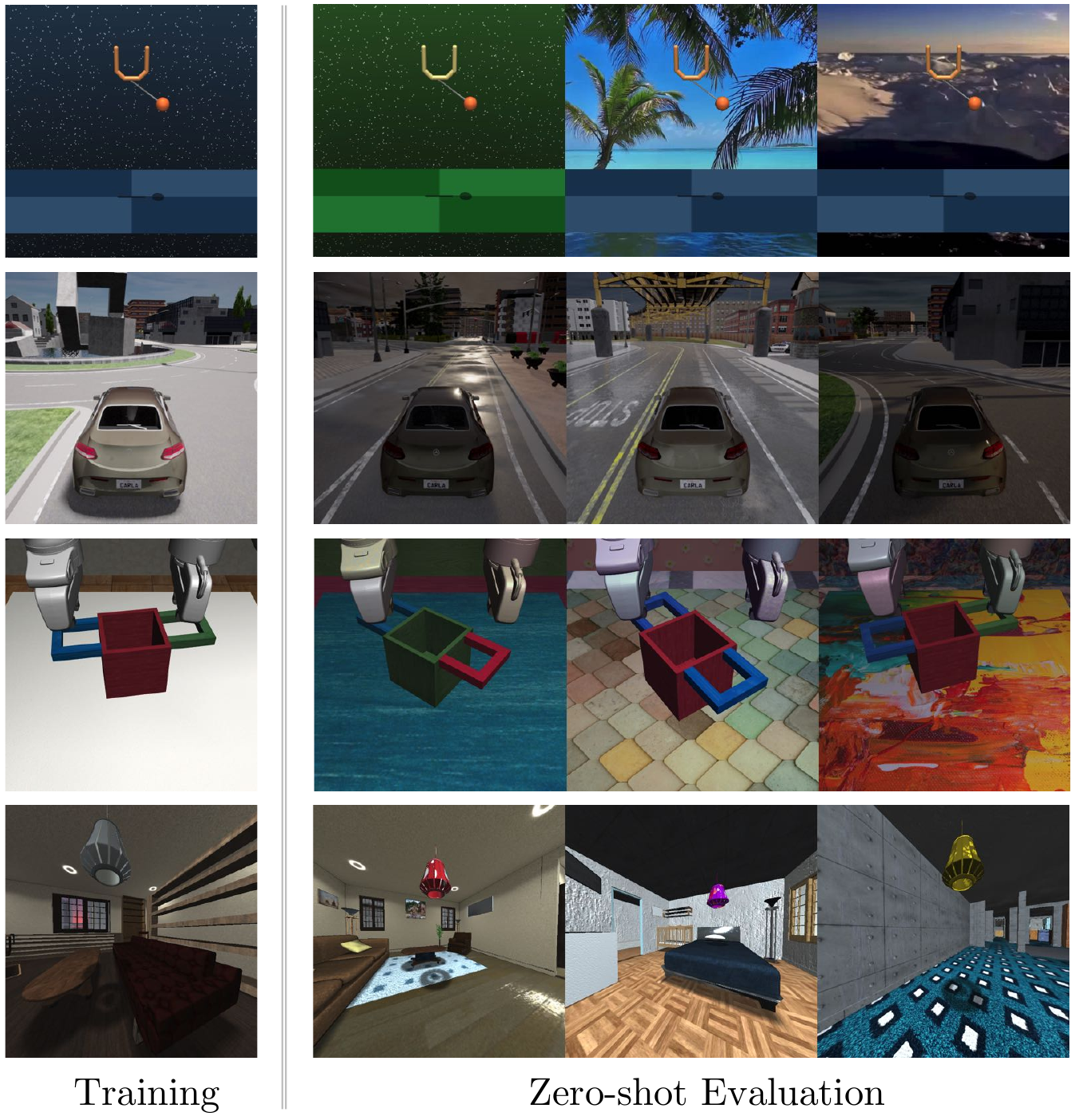}
    \caption{Our proposed benchmark for visual policy generalization in 4 diverse domains.
    Top to bottom: DMControl Suite (15 settings), CARLA autonomous driving (5 weathers), Robosuite (12 settings), and iGibson indoor navigation (20 rooms).}
    \label{fig:benchmark}
    \vspace{-0.13in}
\end{figure}

\section{Introduction}

\begin{figure*}[t!]
    \centering
    \includegraphics[width=0.92\textwidth]{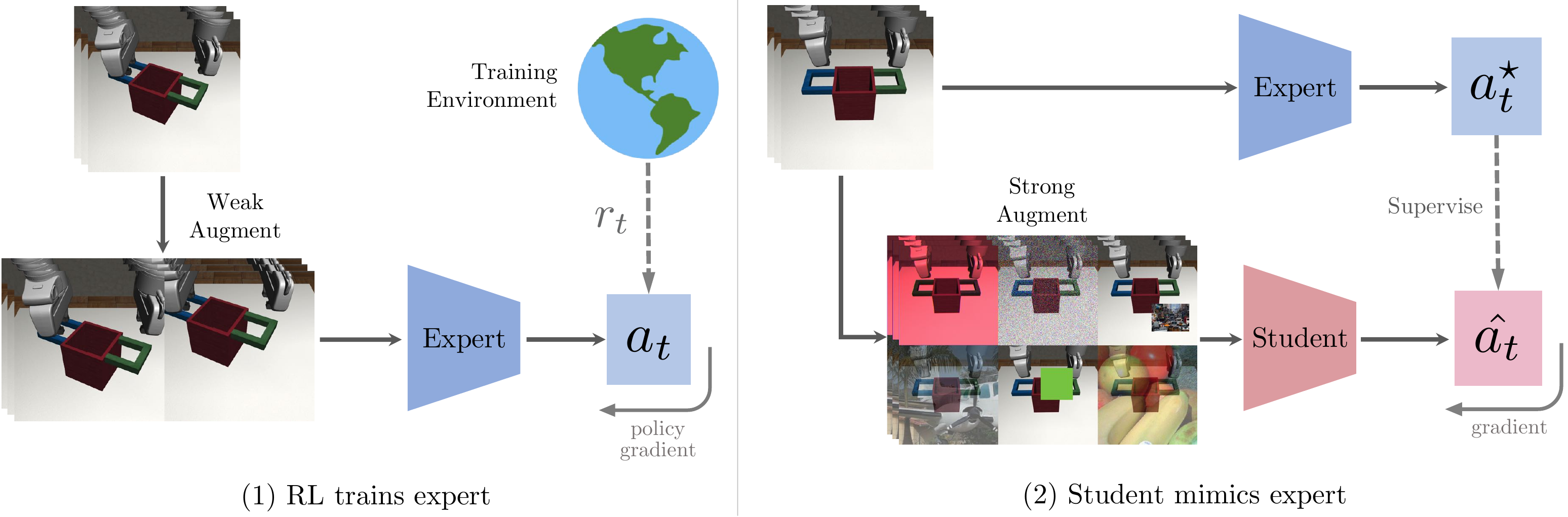}
    \caption{Algorithm overview. \secant training is split into two stages. \textbf{Left, stage 1}: expert policy is trained by RL with weak augmentation (random cropping). \textbf{Right, stage 2}: student receives ground-truth action supervision from the expert at every time step, conditioned on the same observation but with strong augmentations, such as cutout-color, Gaussian noise, Mixup, and Cutmix. The student learns robust visual representations invariant to environment distractions, while maintaining high policy performance.}
    \label{fig:algorithm}
    \vspace{-0.05in}
\end{figure*}

Deep reinforcement learning (RL) from image observations has seen much success in various domains \cite{Mnih2013PlayingAW,levine2016end,andrychowicz2020learning}.  %
However, generalization remains a major obstacle towards reliable deployment. Recent studies have shown that RL agents struggle to generalize to new environments, even with similar tasks \citep{Farebrother2018GeneralizationAR, Gamrian2019TransferLF, cobbe2019leveraging, Song2020ObservationalOI}. %
This suggests that the learned RL policies fail to develop robust representations against irrelevant environmental variations. 

Factors of variation in RL problems can be grouped into three main categories: generalization over different visual appearances \citep{cobbe2018quantifying, Gamrian2019TransferLF, lee2020network}, dynamics \citep{Packer2018AssessingGI}, and environment structures \citep{wang2016learning, Beattie2016DeepMindL, cobbe2019leveraging}. In this work, we mainly focus on zero-shot generalization to unseen environments of different visual appearances, but the same semantics.    

One well-explored solution for better generalization is data augmentation \cite{LeCun1998GradientbasedLA}. For image observations in RL, augmentation can be either manually engineered into the simulator, also known as domain randomization \cite{Tobin2017DomainRF}, or automatic \cite{laskin2020reinforcement}.  
Prior works \cite{berthelot2019remixmatch,sohn2020fixmatch} distinguish between \textit{weak} augmentations like random cropping, and \textit{strong} augmentations that heavily distort the image, such as Mixup \cite{zhang2017mixup} and Cutmix \cite{yun2019cutmix}. Strong augmentations are known to induce robust and generalizable representations for image classification \cite{hendrycks2019augmix}.  
However, naively transplanting them to RL hinders training and results in suboptimal \mbox{performance} \cite{laskin2020reinforcement}. Therefore, weak augmentations like random cropping are the most effective for RL at training time \cite{kostrikov2020image}. This poses a dilemma: more aggressive augmentations are necessary to cultivate better generalization for the visual domain \cite{hendrycks2019augmix}, but RL does not benefit to the same extent as supervised learning since the training is fragile to excessive data \mbox{variations}.

We argue that the dilemma exists because it conflates two problems: policy learning and robust representation learning. 
To decouple them, we draw inspiration from policy distillation \cite{rusu2015policy} where a student policy distills knowledge from one or more experts. The technique is used for different purposes, such as efficient policy deployment, multi-task RL, and policy transfer \cite{teh2017distral,arora2018multi,czarnecki2019distilling}. In this work, we introduce a new instantiation of policy distillation that addresses the dilemma effectively. 

\textbf{Summary of our contributions}:
\vspace{-0.05in}
\begin{itemize}[leftmargin=*]
\item We propose \secant (\textbf{S}elf \textbf{E}xpert \textbf{C}loning for \textbf{A}daptation to \textbf{N}ovel \textbf{T}est-environments), a novel algorithm that solves policy learning and robust representation learning \mbox{\textit{sequentially}}, which achieves strong zero-shot generalization performance to unseen visual environments.
\vspace{-0.02in}
\item We design and standardize a diverse and challenging suite of benchmarks in 4 domains: Deepmind Control Suite (DMControl), autonomous driving, robotic manipulation, and indoor object navigation. Except for DMControl, the other 3 environments feature test-time visual appearance drifts that are representative of real-world applications. 
\vspace{-0.02in}
\item We demonstrate that \secant is able to dominate prior state-of-the-art methods on the majority of tasks, often by substantial margins, across all 4 domains.
\end{itemize}

Our \textbf{key insight} is to solve policy optimization first, and then to robustify its representation by imitation learning with strong augmentations.  
First, an expert neural network is trained by RL with random cropping on the original environment. It learns a high-performance policy but cannot handle distribution shifts. 
Second, a student network learns to mimic the behavior of the expert, but with a crucial difference: the expert computes the ground-truth actions from \textit{unmodified} observations, while the student learns to predict the same actions from heavily \textit{corrupted} observations. 
The student optimizes a supervised learning objective, which has better training stability than RL, and the strong augmentations greatly remedy overfitting at the same time. 
Thus, \secant is able to acquire robust representations without sacrificing policy performance.

Our method is strictly zero-shot, because no reward signal is allowed at test time, and neither the expert nor the student sees the test environments during training. \secant is trained once and does not perform any test-time adaptation. In contrast, PAD \cite{hansen2020self}, a prior SOTA method, adds a self-supervised auxiliary loss on intermediate representations during training, and continues to fine-tune the policy weights using this loss signal at testing. \secant is more efficient compared to PAD, because the latter requires expensive gradient computation at every inference step and is impractical to deploy on mobile robots. %

We benchmark on Deepmind Control Suite (DMControl) with randomized color and video backgrounds \cite{hansen2020self}, and show that \secant is able to outperform prior SOTA in \textbf{14 out of 15} unseen environments with an average score increase of \textbf{26.5\%}. While DMControl is a popular benchmark, its test-time variations are artificial and not representative of real applications. Therefore, we further construct 3 new benchmarks with more realistic distribution shifts (Fig.~\ref{fig:benchmark}), based on existing simulators: \textbf{(1) Robosuite} \cite{zhu2020robosuite}: single-hand and bimanual robotic tasks. We add new appearances of table, background, and objects of interest that are varied at test time; 
\textbf{(2) CARLA} \cite{dosovitskiy2017carla}: autonomous driving across 5 unseen weather conditions that feature highly realistic rendering of raining, sunlight changes, and shadow effects.
\textbf{(3) iGibson} \cite{shen2020igibsonicra}: indoor object navigation in 20 distinct rooms with a large variety of interior design and layouts that we standardize. We hope that these new challenging environments will facilitate more progress towards generalizable visual policy learning.

\section{Related Work}

\para{Generalization in Deep RL.} There is a plethora of literature that highlights the overfitting problem in deep RL \cite{Rajeswaran2017TowardsGA, Packer2018AssessingGI, Zhang2018ADO, Justesen2018IlluminatingGI, Machado2018RevisitingTA, cobbe2018quantifying, Wang2019OnTG, cobbe2019leveraging,  Yarats2019ImprovingSE, Raileanu2020RIDERI}. 
One class of approach is to re-design the training objectives to induce invariant representations directly. \citet{zhang2020learning} and \citet{srinivas2020curl} aim to learn robust features via deep metric learning \cite{ferns2014bisimulation}. 
\citet{rao2020rlcyclegan} combines RL with CycleGAN. 
\citet{jiang2020prioritized} employs automatic curriculum for generalization. 
PAD \cite{hansen2020self} adds a self-supervised auxiliary component that can be adapted at test time. In contrast to these prior works, \secant is a plug-and-play method that neither modifies the existing RL algorithm, nor requires computationally expensive test-time fine-tuning. 
Similar to us, ATC \cite{stooke2020atc} separates representation learning from RL. It pretrains an encoder, fine-tunes with reward, and evaluates in the \textit{same} environment. In contrast, \secant solves policy learning first before robustification, and focuses heavily on zero-shot generalization instead.  

Other works \cite{Farebrother2018GeneralizationAR,cobbe2018quantifying} apply regularization techniques originally developed for supervised learning, such as L2 regularization, BatchNorm~\cite{Ioffe2015BatchNA}, and dropout~\cite{Srivastava2014DropoutAS}. \citet{igl2019selectivenoise} regularizes RL agents via selective noise injection and information bottleneck. These methods improve policy generalization in Atari games and CoinRun~\cite{cobbe2018quantifying}. \secant is orthogonal to these techniques and can be combined for further improvements. We also contribute a new benchmark with more realistic tasks and variations than video games. %

\para{Data augmentation and robustness.} Semantic-preserving image transformations have been widely adopted to improve the performance and robustness of computer vision systems \cite{Hendrycks2018BenchmarkingNN, hendrycks2019augmix, berthelot2019mixmatch, sohn2020fixmatch}. Domain randomization (DR) \cite{Tobin2017DomainRF,Peng_2018} produces randomized textures of environmental components. It is a special type of data augmentation that requires extensive manual engineering and tuning of the simulator \cite{pinto2017asymmetric,yang2019single}. 
RL training, however, benefits the most from weak forms of augmentations that do not add extra difficulty to the policy optimization process \cite{laskin2020reinforcement,kostrikov2020image,raileanu2020automatic,hansen2020self}. By design, \secant unlocks a multitude of strong augmentation operators that are otherwise suboptimal for training in prior works. We successfully employ techniques from supervised image classification like Cutmix \cite{yun2019cutmix} and Mixup \cite{zhang2017mixup}; the latter has also been explored in \cite{wang2020rlmixup}. 

\para{Policy distillation.} \secant belongs to the policy distillation family, a special form of knowledge distillation \cite{hinton2015distilling} for RL. Prior works use policy distillation for different purposes \cite{czarnecki2019distilling}. \citet{chen2020cheat} and \citet{lee2020terrain} train an expert with privileged simulator information (e.g. groundtruth physical states) to supervise a student policy that can only access limited sensors at deployment. \citet{zhou2020domain} transfers navigation policies across domains through an intermediate proxy model. \citet{Igl2020TheIO} reduces the non-stationary effects of RL environment by repeated knowledge transfer. Other works involve multi-task student networks \cite{rusu2015policy,teh2017distral,arora2018multi} that distill from multiple experts simultaneously. \secant differs from these works because our expert and student share the same task and observation information, but shoulder different responsibilities: expert handles policy optimization while student addresses visual generalization. 

Our method is related to FixMatch \cite{sohn2020fixmatch}, which imposes a pseudo-label distillation loss on two different augmentations of the same image. In contrast, our expert only needs to overfit to the training environment, while the student distills from a frozen expert to learn robust representation. Concurrent work \citet{hansen2020soda} also validates the benefit of decoupling and strong augmentation. In comparison, \secant is conceptually simpler and does not require modifying the RL training pipeline.
Another closely related field is imitation learning \cite{schaal1997LFD,argall2009imitationsurvey,ross2011dagger,ho2016gail}. Our student imitates without external demonstration data, hence the name ``self-expert cloning".

\section{Preliminaries}
\label{sec:preliminaries}

\para{Soft Actor-Critic.}
In this work, we mainly consider continuous control from raw pixels. The agent receives an image observation $o\in \mathbb{R}^{C\times H\times W}$ and outputs a continuous action $a \in \mathbb{R}^d$.
SAC~\cite{haarnoja2018soft,haarnoja2018soft2} is a state-of-the-art off-policy RL algorithm. 
It learns a policy $\pi(a|o)$ and a critic $Q(o,a)$ that maximize a weighted combination of reward and policy entropy, $\mathbb{E}_{(o_t, a_t) \sim \pi} \left[\sum_t r_t + \alpha \mathcal{H}(\pi(\cdot|o_t))\right]$. SAC stores experiences into a replay buffer $\gD$. The critic parameters are updated by minimizing the Bellman error using transitions sampled from $\mathcal{D}$:
\begin{equation}
    \label{eq:SAC_critic}
    \gL_Q = \mathbb{E}_{(o_t, a_t)\sim\mathcal{D}}\left[\big(Q(o_t,a_t) - (r_t + \gamma V(o_{t+1}))\big)^2\right]
\end{equation}
By sampling an action under the current policy, we can estimate the soft state value as following:
\begin{equation}
    V(o_{t+1})=\mathbb{E}_{a'\sim\pi}\left[{\bar{Q}}(o_{t+1},a')-\alpha \log\pi(a'|o_{t+1})\right]
\end{equation}
where ${\bar{Q}}$ denotes an exponential moving average of the critic network. The policy is updated by minimizing the divergence from the exponential of the soft-Q function:
\begin{equation}
    \label{eq:SAC_policy}
    \gL_\pi =-\mathbb{E}_{a_t\sim\pi}\left[Q(o_t,a_t)-\alpha \log\pi(a_t|o_t)\right]
\end{equation}
where $\alpha$ is a learnable temperature parameter that controls the stochasticity of the optimal policy.

\para{Dataset Aggregation (DAgger).}
\citet{ross2011dagger} is an iterative imitation learning algorithm with strong performance guarantees. %
First, it rolls out an expert policy $\pi_e$ to seed an experience dataset $\gD ^ 0$. The student policy $\pi_s ^ 0$ is trained by supervised learning to best mimic the expert on those trajectories. Then at iteration $i$, it rolls out $\pi_s ^ i$ to collect more trajectories that will be added to $\gD ^ i$. $\pi_s ^ {i+1}$ will then be trained on the new \textit{aggregated} dataset $\gD^ {i+1}$, and the process repeats until convergence. Even though more advanced imitation algorithms have been developed \cite{ho2016gail}, DAgger is conceptually simple and well-suited for \secant because our student network can query the expert for dense supervision at every time step.

\newcommand{\addDmcFig}[1]{\includegraphics[width=\linewidth]{figs/dmc-env/#1.png}}

\begin{table*}[t]
\vskip -0.05in
\caption{DMControl: \secant outperforms prior SOTA methods substantially in \textbf{14 out of 15} settings with \textbf{+26.5\%} boost on average.}
\label{table:dmc-core}
\vskip 0.1in
\centering
\resizebox{\textwidth}{!}{%
\begin{tabular}{C{0.1\textwidth}|r|ccccccc}

\toprule
Setting & Task & \secant\, (Ours) & SAC & SAC+crop & DR & NetRand & SAC+IDM & PAD \\ \midrule 

\multirow{8}{*}{\addDmcFig{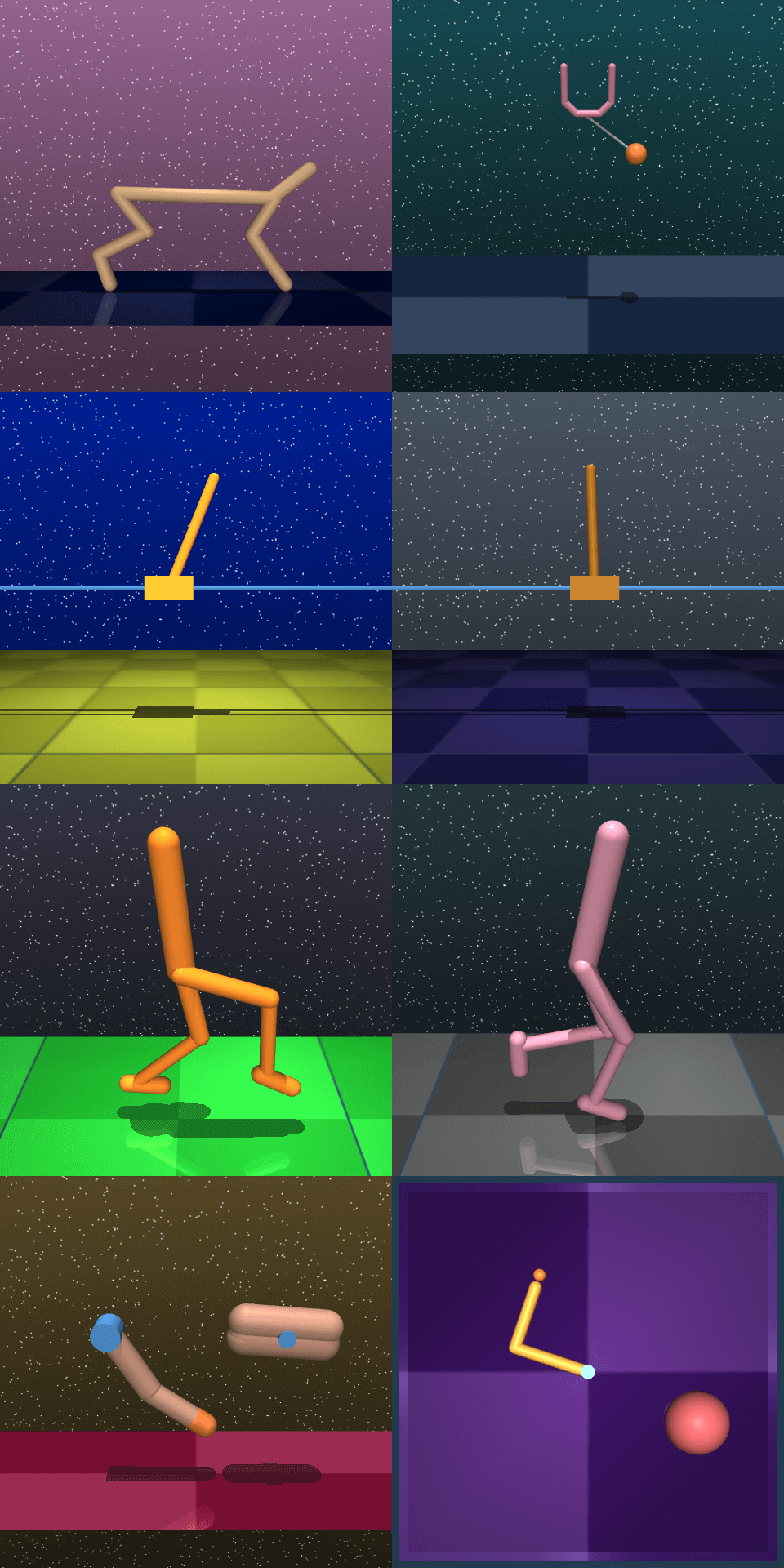}} & Cheetah run & $\bestscore{582\pm64\hphantom{0}}$ \,(\bestpercent{+88.3}) & $133\pm26\hphantom{0}$ & $100\pm27\hphantom{0}$ & $145\pm29\hphantom{0}$ & $309\pm66\hphantom{0}$ & $121\pm38\hphantom{0}$ & $159\pm28\hphantom{0}$ \\
 & Ball in cup catch & $\bestscore{958\pm7\hphantom{00}}$ \,(\bestpercent{+\hphantom{0}8.1}) & $151\pm36\hphantom{0}$ & $359\pm76\hphantom{0}$ & $470\pm252$ & $886\pm57\hphantom{0}$ & $471\pm75\hphantom{0}$ & $563\pm50\hphantom{0}$ \\
 & Cartpole swingup & $\bestscore{866\pm15\hphantom{0}}$ \,(\bestpercent{+27.2}) & $248\pm24\hphantom{0}$ & $537\pm98\hphantom{0}$ & $647\pm48\hphantom{0}$ & $681\pm122$ & $585\pm73\hphantom{0}$ & $630\pm63\hphantom{0}$ \\
 & Cartpole balance & $\bestscore{992\pm6\hphantom{00}}$ \,(\bestpercent{+\hphantom{0}0.8}) & $930\pm36\hphantom{0}$ & $769\pm63\hphantom{0}$ & $867\pm37\hphantom{0}$ & $984\pm13\hphantom{0}$ & $835\pm40\hphantom{0}$ & $848\pm29\hphantom{0}$ \\
 & Walker walk & $\bestscore{856\pm31\hphantom{0}}$ \,(\bestpercent{+27.6}) & $144\pm19\hphantom{0}$ & $191\pm33\hphantom{0}$ & $594\pm104$ & $671\pm69\hphantom{0}$ & $406\pm29\hphantom{0}$ & $468\pm47\hphantom{0}$ \\
 & Walker stand & $\bestscore{939\pm7\hphantom{00}}$ \,(\bestpercent{+\hphantom{0}4.3}) & $365\pm79\hphantom{0}$ & $748\pm60\hphantom{0}$ & $715\pm96\hphantom{0}$ & $900\pm75\hphantom{0}$ & $743\pm37\hphantom{0}$ & $797\pm46\hphantom{0}$ \\
 & Finger spin & $\bestscore{910\pm115}$ \,(\bestpercent{+\hphantom{0}3.1}) & $504\pm114$ & $847\pm116$ & $465\pm314$ & $883\pm156$ & $757\pm62\hphantom{0}$ & $803\pm72\hphantom{0}$ \\
 & Reacher easy & $\bestscore{639\pm63\hphantom{0}}$ \,(\bestpercent{+29.1}) & $185\pm70\hphantom{0}$ & $231\pm79\hphantom{0}$ & $105\pm37\hphantom{0}$ & $495\pm101$ & $201\pm32\hphantom{0}$ & $214\pm44\hphantom{0}$ \\
\midrule 

\multirow{7}{*}{\addDmcFig{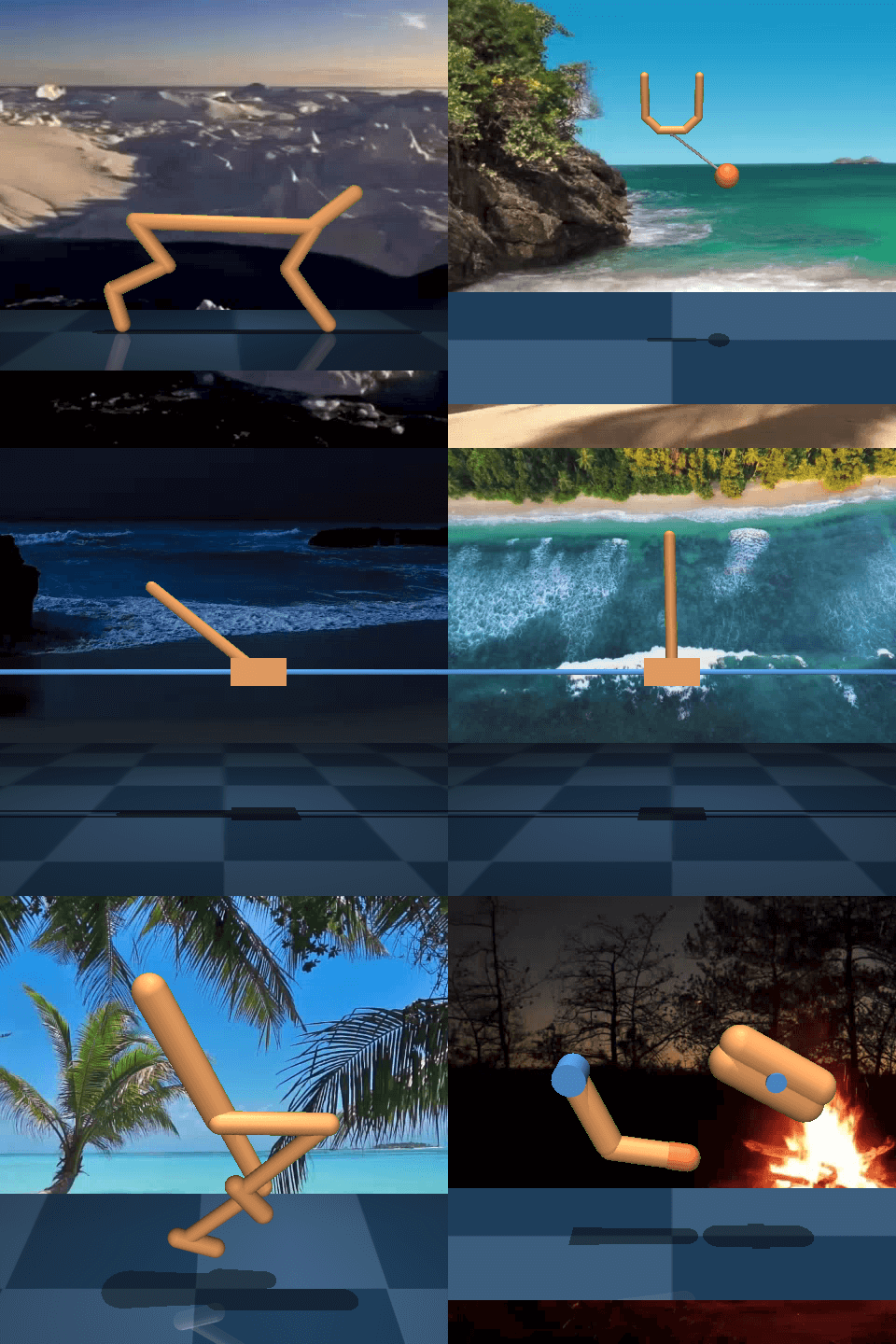}} & Cheetah run & $\bestscore{428\pm70\hphantom{0}}$ \,(\bestpercent{+56.8}) & $\hphantom{0}80\pm19\hphantom{0}$ & $102\pm30\hphantom{0}$ & $150\pm34\hphantom{0}$ & $273\pm26\hphantom{0}$ & $164\pm42\hphantom{0}$ & $206\pm34\hphantom{0}$ \\
 & Ball in cup catch & $\bestscore{903\pm49\hphantom{0}}$ \,(\bestpercent{+57.3}) & $172\pm46\hphantom{0}$ & $477\pm40\hphantom{0}$ & $271\pm189$ & $574\pm82\hphantom{0}$ & $362\pm69\hphantom{0}$ & $436\pm55\hphantom{0}$ \\
 & Cartpole swingup & $\bestscore{752\pm38\hphantom{0}}$ \,(\bestpercent{+44.3}) & $204\pm20\hphantom{0}$ & $442\pm74\hphantom{0}$ & $485\pm67\hphantom{0}$ & $445\pm50\hphantom{0}$ & $487\pm90\hphantom{0}$ & $521\pm76\hphantom{0}$ \\
 & Cartpole balance & $\bestscore{863\pm32\hphantom{0}}$ \,(\bestpercent{+12.7}) & $569\pm79\hphantom{0}$ & $641\pm37\hphantom{0}$ & $766\pm92\hphantom{0}$ & $708\pm28\hphantom{0}$ & $691\pm76\hphantom{0}$ & $687\pm58\hphantom{0}$ \\
 & Walker walk & $\bestscore{842\pm47\hphantom{0}}$ \,(\bestpercent{+17.4}) & $104\pm14\hphantom{0}$ & $244\pm83\hphantom{0}$ & $655\pm55\hphantom{0}$ & $503\pm55\hphantom{0}$ & $694\pm85\hphantom{0}$ & $717\pm79\hphantom{0}$ \\
 & Walker stand & $932\pm15\hphantom{0}$ \,\hphantom{(+00.0\%)} & $274\pm39\hphantom{0}$ & $601\pm36\hphantom{0}$ & $869\pm60\hphantom{0}$ & $769\pm78\hphantom{0}$ & $902\pm51\hphantom{0}$ & $\bestscore{935\pm20\hphantom{0}}$ \\
 & Finger spin & $\bestscore{861\pm102}$ \,(\bestpercent{+21.6}) & $276\pm81\hphantom{0}$ & $425\pm69\hphantom{0}$ & $338\pm207$ & $708\pm170$ & $605\pm61\hphantom{0}$ & $691\pm80\hphantom{0}$ \\

\bottomrule

\end{tabular}
}
\vskip -0.1in
\end{table*}

\newcommand{\addAugFig}[1]{\includegraphics[width=\linewidth]{figs/augs/#1.png}}

\begin{table*}[t]
\caption{Ablation on student augmentations: given the same experts trained with random cropping, we ablate 6 strong augmentations and their mixtures for the student. Combo[1-3] randomly select an augmentation from their pool to apply to each observation. %
}
\label{table:dmc-all-augs}
\vskip 0.15in
\centering
\scriptsize
\setlength{\tabcolsep}{0.3em}
\begin{tabular}{C{0.05\textwidth}|R{0.11\textwidth}|C{0.08\textwidth}C{0.08\textwidth}C{0.08\textwidth}C{0.08\textwidth}C{0.08\textwidth}C{0.08\textwidth}C{0.08\textwidth}C{0.08\textwidth}C{0.08\textwidth}}

\toprule

Setting & Tasks & Combo1 & Combo2 & Combo3 & Cutout-color & Conv & Mixup & Cutmix & Gaussian & Impulse \\ \midrule 

\multirow{3}{*}{\makecell{Random \\ Color}} & Cartpole swingup & $\bestscore{866\pm15}$ & $865\pm17$ & $863\pm15$ & $776\pm33$ & $860\pm15$ & $825\pm16$ & $751\pm43$ & $720\pm86$ & $751\pm45$ \\
 & Cheetah run & $\bestscore{582\pm64}$ & $522\pm166$ & $570\pm50$ & $343\pm153$ & $318\pm123$ & $222\pm38$ & $303\pm82$ & $373\pm110$ & $382\pm121$ \\
 & Walker walk & $856\pm31$ & $854\pm27$ & $832\pm45$ & $701\pm75$ & $\bestscore{866\pm22}$ & $756\pm46$ & $658\pm54$ & $770\pm56$ & $727\pm47$ \\
\midrule 

\multirow{3}{*}{\makecell{Random \\ Video}} & Cartpole swingup & $752\pm38$ & $765\pm55$ & $\bestscore{778\pm37}$ & $607\pm31$ & $556\pm61$ & $677\pm43$ & $647\pm56$ & $580\pm61$ & $639\pm22$ \\
 & Cheetah run & $\bestscore{428\pm70}$ & $409\pm31$ & $406\pm33$ & $183\pm46$ & $229\pm30$ & $309\pm65$ & $209\pm43$ & $196\pm38$ & $224\pm25$ \\
 & Walker walk & $\bestscore{842\pm47}$ & $836\pm36$ & $792\pm59$ & $631\pm52$ & $531\pm54$ & $759\pm64$ & $675\pm22$ & $488\pm31$ & $471\pm17$ \\
 
\midrule 

\multicolumn{2}{c|}{\makecell{Example \\ Augmentations }} & \addAugFig{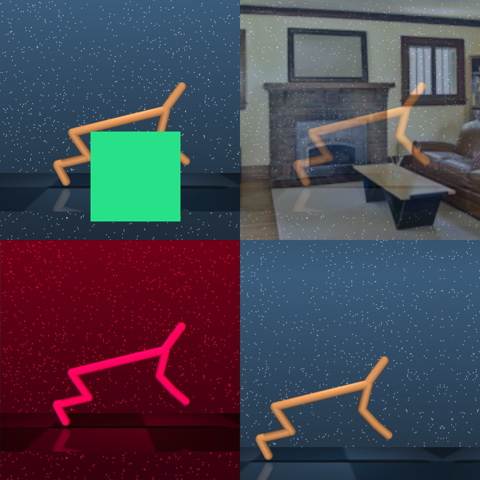} & \addAugFig{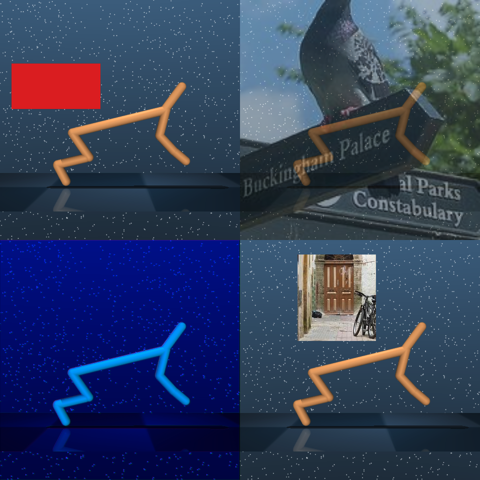} & \addAugFig{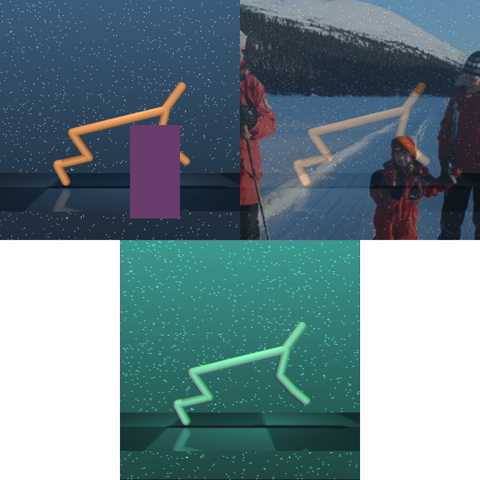} & \addAugFig{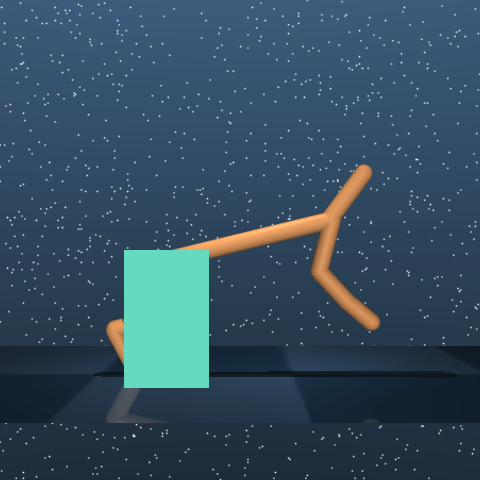} & \addAugFig{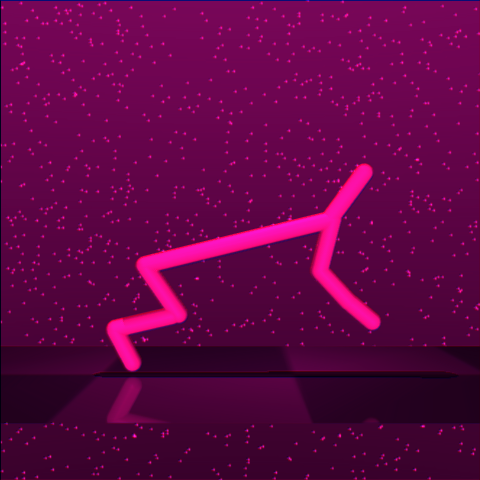} & \addAugFig{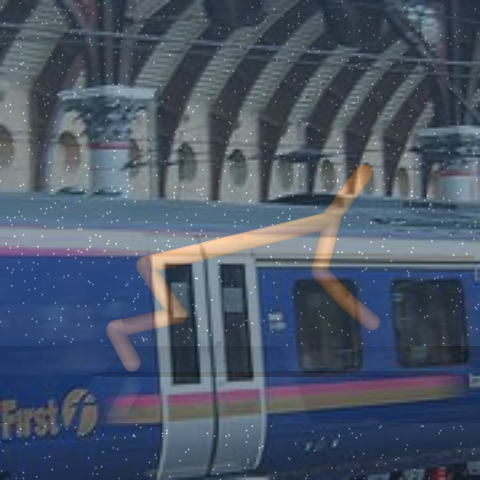} & \addAugFig{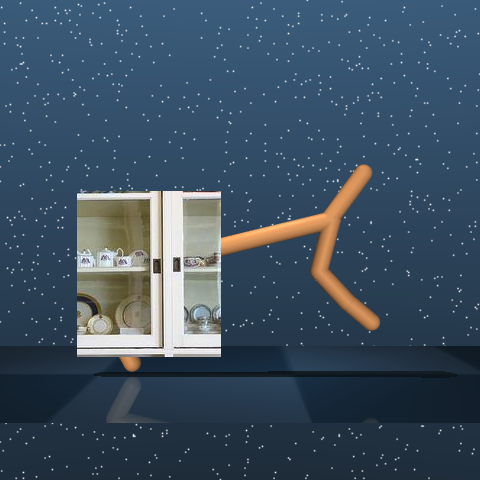} & \addAugFig{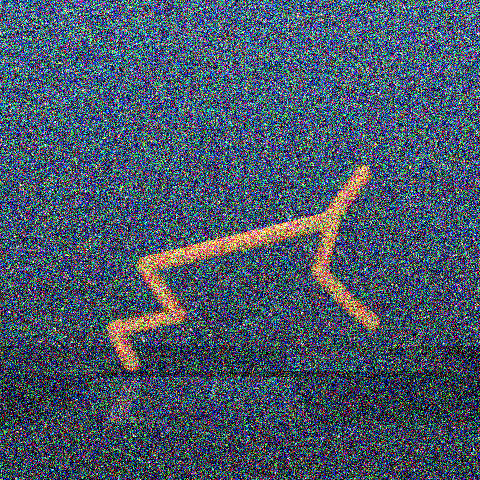} & \addAugFig{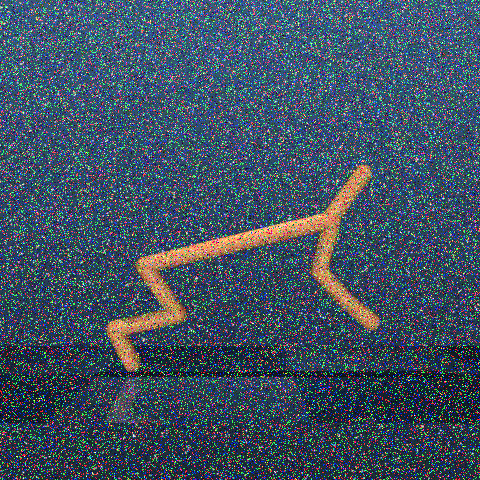} \\

\bottomrule

\end{tabular}

\end{table*}

\section{\secant}

The goal of our proposed self-expert cloning technique is to learn a robust policy that can generalize zero-shot to unseen visual variations. \secant training can be decomposed into two stages.  Algorithm \ref{alg:secant} shows the full pseudocode.

\subsection{Expert policy}

\setlength{\textfloatsep}{12pt}  %
\begin{algorithm}[tb]
\caption{\secant: Self-Expert Cloning}
\label{alg:secant}
\begin{algorithmic}[1]
\STATE $\pi_e$, $\pi_s$: randomly initialized expert and student policies 
\STATE $\gF_{weak}$, $\gF_{strong}$: sets of image augmentations
\STATE $\gB$: experience replay buffer

\FOR{$t$ in $1, \ldots, T_{RL}$}
    \STATE Sample experience batch $\tau_t = (o_t, a_t, o_{t+1}, r) \sim \gB$
    \STATE Sample weak augmentation $f \sim \gF_{weak}$
    \STATE Augment $o_t = f(o_t); o_{t+1} = f(o_{t+1}) $
    \STATE Update $\pi_{e}$ to minimize $\gL_{RL}(\tau_t)$
\ENDFOR
\STATE Roll out $\pi_{e}$ to collect an initial dataset $\gD$ of trajectories  %

\FOR{$t$ in $1, \ldots, T_{imitate}$}
    \STATE Sample observation batch $o \sim \gD$
    \STATE Sample strong augmentation $f \sim \gF_{strong}$
    \STATE Update $\pi_{s}$ to minimize $\| \pi_{s}(f(o)) - \pi_{e}(o) \|_F$ 
    \STATE Roll out $\pi_{s}$ for one environment step and add to the dataset $\gD \leftarrow \gD \cup \{ o_s \} $
\ENDFOR
    
\end{algorithmic}
\end{algorithm}

In the first stage, we train a high-performance expert policy in the original environment with weak augmentations. In visual continuous control tasks, the policy is parametrized by a feed-forward deep convolutional network $\pi_e(\gO; \theta_e): \mathbb{R}^{C \times H \times W} \rightarrow \mathbb{R}^{d}$ that maps an image observation to a $d$-dimensional continuous action vector. In practice, we employ a frame stacking technique that concatenates $T$ consecutive image observations along the channel dimension to incorporate temporal information \cite{Mnih2013PlayingAW}.  
The augmentation operator is a semantic-preserving image transformation $f : \mathbb{R}^{C \times H \times W} \rightarrow \mathbb{R}^{C' \times H' \times W'}$.  
Prior works have found that random cropping performs the best in a range of environments, therefore we adopt it as the default weak augmentation for the expert \cite{laskin2020reinforcement}.

The expert can be optimized by any standard RL algorithm. We select Soft Actor-Critic (SAC) due to its wide adoption in continuous control tasks \cite{haarnoja2018soft,haarnoja2018soft2}. The expert is optimized by gradient descent to minimize the SAC objectives (Equations \ref{eq:SAC_critic} and \ref{eq:SAC_policy}). 
Since we place little restrictions on the expert, our method can even be used to robustify pre-trained policy network checkpoints, such as the RL Model Zoo \cite{antonin2018rlzoo}.

\subsection{Student policy distillation}

In the second stage, we train a student network to predict the optimal actions taken by the expert, conditioned on the same observation but with heavy image corruption. This stage does \textit{not} need further access to the reward signal. Formally, the student is also a deep convolutional network $\pi_s(\gO; \theta_s): \mathbb{R}^{C \times H \times W} \rightarrow \mathbb{R}^{d}$ that may have different architecture from the expert. The student policy distills from the expert following the DAgger imitation procedure (Sec \ref{sec:preliminaries}). First, we roll out the \textit{expert} policy to collect an initial dataset $\gD$ of trajectories. Next, at each iteration, we select a strong augmentation operator $f \sim \gF_{strong}$ and apply it to a batch of observations $o$ sampled from $\gD$. We alternate between (1) updating the student's parameters by gradient descent on a supervised regression loss: $\mathcal{L}(o; \theta_s) = \| \pi_{s}(f(o)) - \pi_{e}(o) \|_F$ and (2) adding more experiences to $\gD$ under the latest \textit{student} policy.

In the experiments, we consider 1 type of weak augmentation (random cropping) and 6 types of strong augmentation techniques developed in RL and robust image classification literature \cite{Hendrycks2018BenchmarkingNN,hendrycks2019augmix,laskin2020reinforcement,lee2020network}.  
We refer to weak augmentations as the ones that can substantially improve RL optimization at training time, while strong augmentations are less effective as they make training more difficult. We focus only on random cropping for weak augmentation in this work, and defer other potential operators to future works. Below are brief descriptions of the augmentations we study:

\textbf{Cutout-color (\texttt{Cc})}: inserts a small rectangular patch of random color into the observation at a random position.
\textbf{Random convolution (\texttt{Cv})}: passes the input observation through a random convolutional layer.
\textbf{Gaussian (\texttt{G})}: adds Gaussian noise.
\textbf{Impulse (\texttt{I})}: adds the color analogue of salt-and-pepper noise.
\textbf{Mixup (\texttt{M})} \cite{zhang2017mixup}: linearly blends the observation with a distracting image $I$: $f(o) = \alpha o + (1 - \alpha) I$. We randomly sample $I$ from 50K COCO images \cite{lin2014microsoftcoco}, and sample $\alpha \sim \text{Uniform}(0.2, 0.6)$.
\textbf{Cutmix (\texttt{Cm})} \cite{yun2019cutmix}: similar to Cutout-color except that the patch is randomly sampled from COCO images.
These augmentations can be categorized into low-frequency noise (\texttt{\textbf{Cc}} and \texttt{\textbf{Cv}}), %
high-frequency unstructured noise (\texttt{\textbf{G}} and \texttt{\textbf{I}}), %
and high-frequency structured noise (\texttt{\textbf{M}} and \texttt{\textbf{Cm}}).  %
Mixup and Cutmix with image distractions are novel operators that have not been studied for visual policy generalization.

We also investigate combinations of the above, and find empirically that random sampling from low-frequency and high-frequency \textit{structured} noise types yields the best overall results. We note that adding random cropping to the mix benefits performance slightly, likely because it improves the spatial invariance of the student's representation. We design three combination recipes (Table \ref{table:dmc-all-augs}): Combo1 (\texttt{Cc+Cv+M+Crop}), Combo2 (\texttt{Cc+Cv+M+Cm+Crop}), and Combo3 (\texttt{Cc+Cv+M}). We have not done an exhaustive search, so it is possible that better combinations exist.

\newcommand{\addRoboFig}[1]{\includegraphics[width=\linewidth]{figs/robo-env/#1.png}}

\begin{table*}[t!]
\caption{Robosuite results. The 3 sets of test environments are progressively harder (\textit{easy}, \textit{hard}, and \textit{extreme}) with more distracting textures of the table, floor, and objects. \secant gains an average of +337.8\% reward over prior SOTA.}
\label{table:robo-core}
\vskip 0.1in
\centering
\resizebox{\textwidth}{!}{%
\begin{tabular}{C{0.1\textwidth}|r|ccccccc}

\toprule
Setting & Tasks & \secant\, (Ours) & SAC & SAC+crop & DR & NetRand & SAC+IDM & PAD \\ \midrule 

\multirow{4}{*}{\addRoboFig{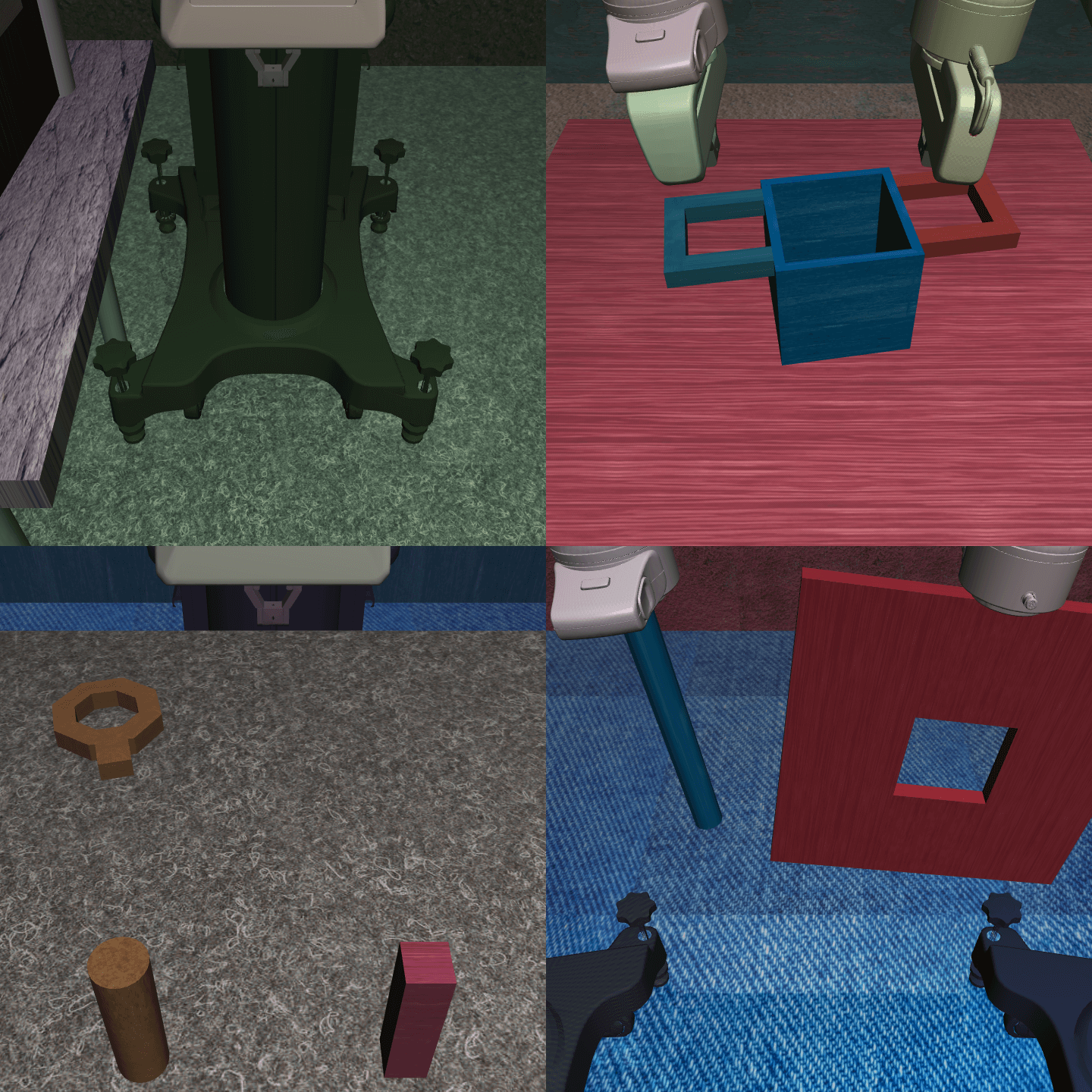}} & Door opening & $\bestscore{782\pm93\hphantom{0}}$ \,(\bestpercent{+\hphantom{0}78.5}) & $17\pm12$ & $10\pm8$ & $177\pm163$ & $438\pm157$ & $3\pm2$ & $2\pm1$ \\
 & Nut assembly & $\bestscore{419\pm63\hphantom{0}}$ \,(\bestpercent{+\hphantom{0}73.1}) & $3\pm2$ & $6\pm5$ & $12\pm7$ & $242\pm28$ & $13\pm12$ & $11\pm10$ \\
 & Two-arm lifting & $\bestscore{610\pm28\hphantom{0}}$ \,(\bestpercent{+883.9}) & $29\pm11$ & $23\pm10$ & $41\pm9$ & $62\pm43$ & $20\pm8$ & $22\pm7$ \\
 & Peg-in-hole & $\bestscore{837\pm42\hphantom{0}}$ \,(\bestpercent{+114.6}) & $186\pm62$ & $134\pm72$ & $139\pm37$ & $390\pm68$ & $150\pm41$ & $142\pm37$ \\
\midrule 

\multirow{4}{*}{\addRoboFig{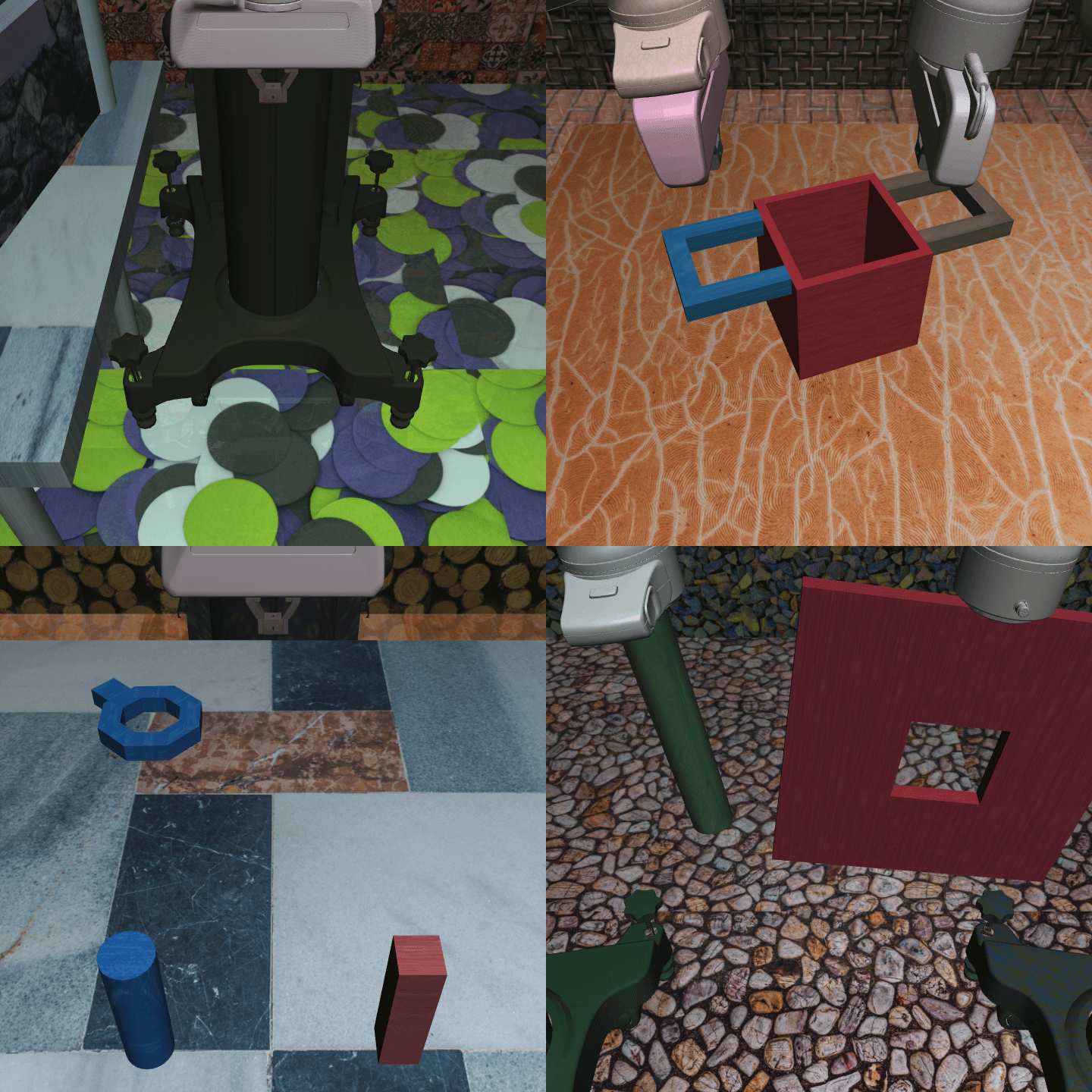}} & Door opening & $\bestscore{522\pm131}$ \,(\bestpercent{+292.5}) & $11\pm10$ & $11\pm7$ & $37\pm31$ & $133\pm82$ & $2\pm1$ & $2\pm1$ \\
 & Nut assembly & $\bestscore{437\pm102}$ \,(\bestpercent{+141.4}) & $6\pm7$ & $9\pm8$ & $33\pm18$ & $181\pm53$ & $34\pm28$ & $24\pm26$ \\
 & Two-arm lifting & $\bestscore{624\pm40\hphantom{0}}$ \,(\bestpercent{+923.0}) & $28\pm11$ & $27\pm9$ & $61\pm15$ & $41\pm25$ & $17\pm6$ & $19\pm8$ \\
 & Peg-in-hole & $\bestscore{774\pm76\hphantom{0}}$ \,(\bestpercent{+140.4}) & $204\pm81$ & $143\pm62$ & $194\pm41$ & $322\pm72$ & $165\pm75$ & $164\pm69$ \\
\midrule 

\multirow{4}{*}{\addRoboFig{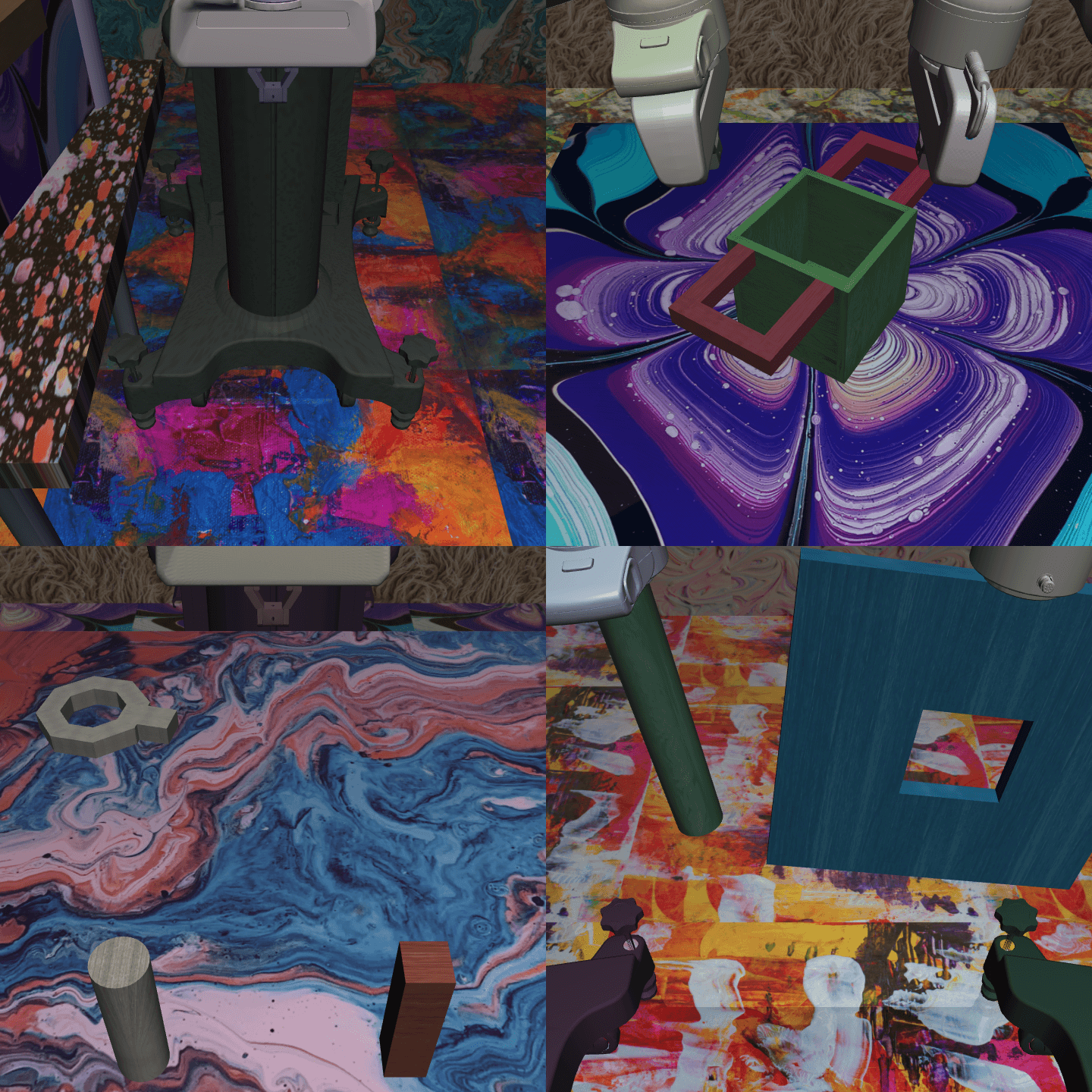}} & Door opening & $\bestscore{309\pm147}$ \,(\bestpercent{+120.7}) & $11\pm10$ & $6\pm4$ & $52\pm46$ & $140\pm107$ & $2\pm1$ & $2\pm1$ \\
 & Nut assembly & $\bestscore{138\pm56\hphantom{0}}$ \,(\bestpercent{+\hphantom{0}53.3}) & $2\pm1$ & $10\pm7$ & $12\pm7$ & $90\pm61$ & $4\pm3$ & $4\pm3$ \\
 & Two-arm lifting & $\bestscore{377\pm37}$ \,(\bestpercent{+1156.7}) & $25\pm7$ & $12\pm6$ & $30\pm13$ & $12\pm11$ & $24\pm10$ & $21\pm10$ \\
 & Peg-in-hole & $\bestscore{520\pm47\hphantom{0}}$ \,(\bestpercent{+\hphantom{0}75.7}) & $164\pm63$ & $130\pm81$ & $154\pm34$ & $296\pm90$ & $155\pm73$ & $154\pm72$ \\

\bottomrule

\end{tabular}
}
\end{table*}

\section{Experiments}
\label{sec:experiments}

We propose a new benchmark of 4 diverse domains (Fig. \ref{fig:benchmark}) to systematically assess the generalization ability of visual agents. They offer a wide spectrum of visual distribution shifts for testing. In each domain, we investigate how well an algorithm trained in one environment performs on various unseen environments in a zero-shot setting, which disallows reward signal and extra trials at test time.

For each task, we benchmark \secant extensively against prior state-of-the-art algorithms: 
\textbf{SAC}: plain SAC with no augmentation. 
\textbf{SAC+crop}: SAC with time-consistent random cropping \cite{kostrikov2020image}.
\textbf{DR}: domain randomization. To simulate realistic deployment, our randomized training distributions are narrower than the test distributions. 
\textbf{NetRand}: Network Randomization \cite{lee2020network}, which augments the observation image by random convolution.
\textbf{SAC+IDM}: SAC trained with an auxiliary inverse dynamics loss \cite{pathakICMl17curiosity}. 
\textbf{PAD}: prior SOTA method on top of SAC+IDM that fine-tunes the auxiliary head at test time \cite{hansen2020self}.  
Following prior works \cite{hansen2020self} on DMControl, we repeat training across 10 random seeds to report the mean and standard deviation of the rewards. We use 5 random seeds for all other simulators and ablation studies. 

\para{Algorithm details.}  \secant builds upon SAC, and adopts similar hyperparameters and network architecture as \citet{kostrikov2020image}. 
Observations are stacks of 3 consecutive RGB frames.
For all tasks, we use a 4-layer feed-forward ConvNet with no residual connection as encoder for both the \secant expert and student, although they do not have to be identical. PAD, however, requires a deeper encoder network (11 layers) to perform well in DMControl \cite{hansen2020self}. For all other simulators, we conduct a small grid search and find that 6-layer encoders work best for both SAC+IDM and PAD. 
After the encoder, 3 additional fully connected layers map the visual embedding to action. 
We include a detailed account of all hyperparameters and architecture in Appendix \ref{appendix:algorithm-hypers}.

\begin{figure}[t!] 
\tiny
\centering

\setlength\tabcolsep{3pt}
\begin{tabularx}{\linewidth}{ c c }

Cheetah run & Walker walk \\ 

\includegraphics[width=0.48\linewidth]{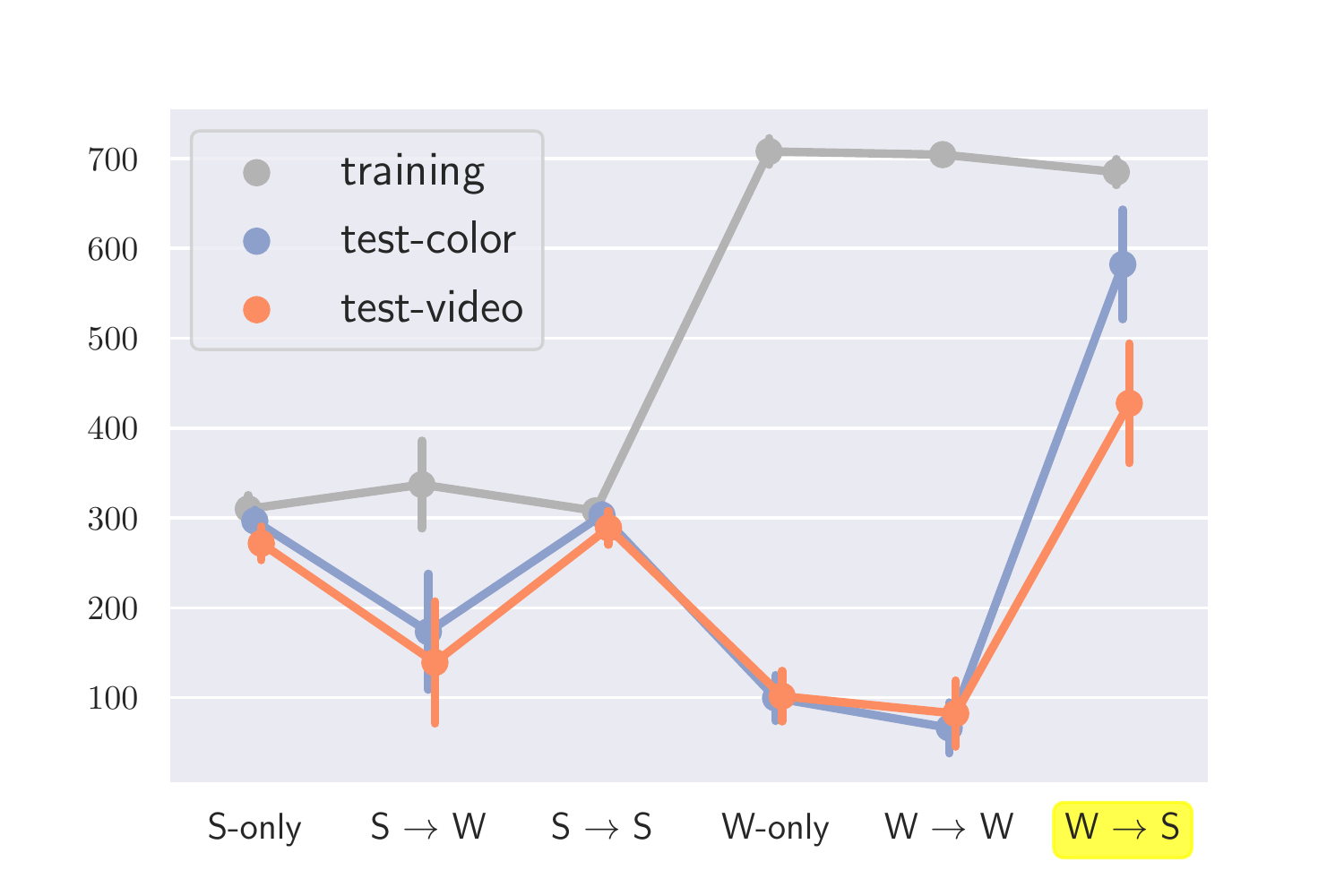}  &  
\includegraphics[width=0.48\linewidth]{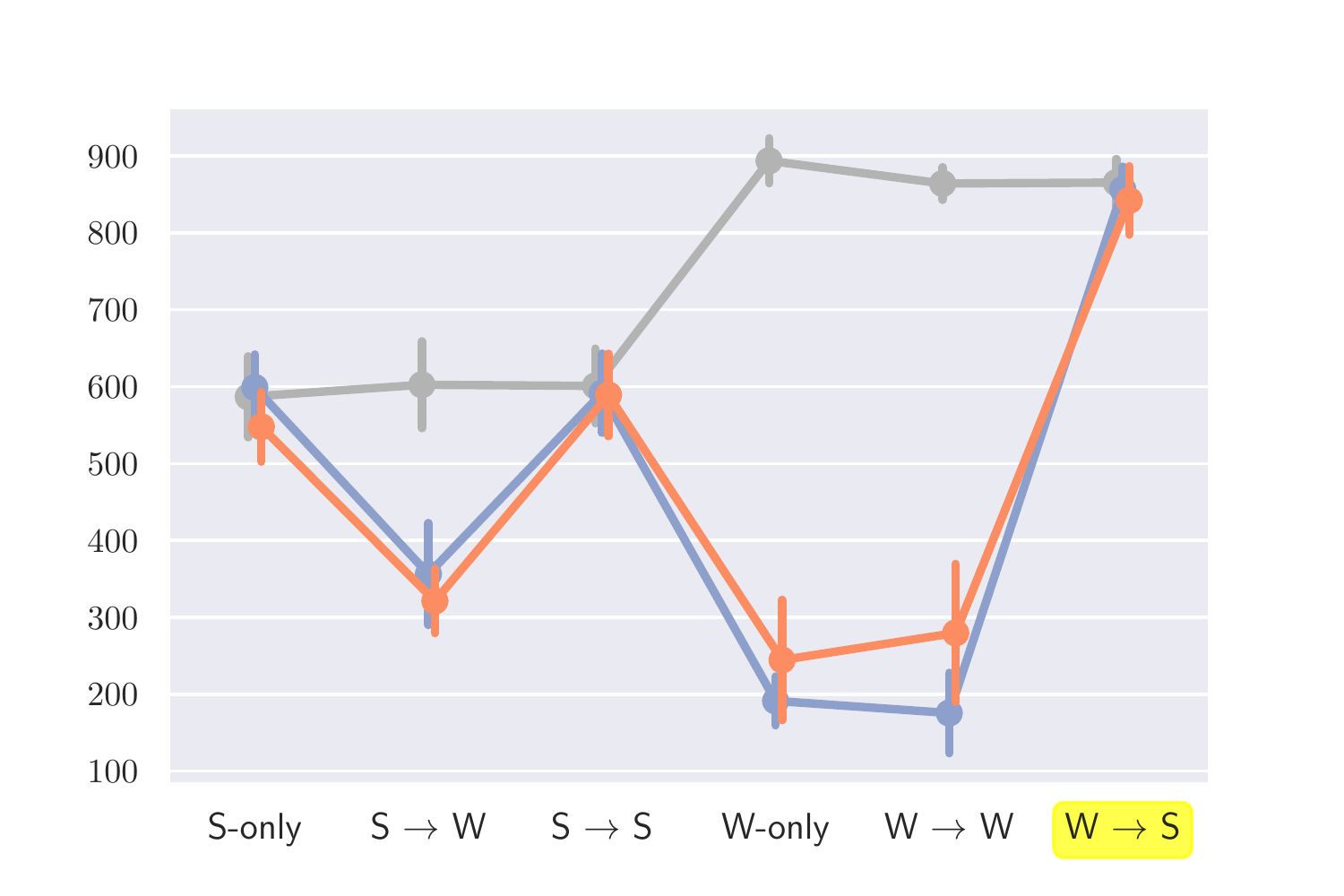} 
\end{tabularx}

\caption{Ablation on different strategies to apply augmentation. ``S-only" denotes single-stage policy trained with strong augmentation, and S $\rightarrow$ W means strongly-augmented expert imitated by weakly-augmented student. The recipe for \secant is \narrowcolorbox{yellow}{W $\rightarrow$ S}. 
}

\label{fig:dmc-weak-strong}
\end{figure}

\subsection{Deepmind Control Suite}

We follow the settings in \citet{hansen2020self} and experiment with 8 tasks from DMControl. We measure generalization to (1) randomized colors of the background and robot itself, and (2) natural videos as dynamic background (Fig. \ref{fig:benchmark}). \secant \textit{significantly outperforms} prior SOTA in all but one task, often by substantial margins up to \textbf{88.3\%} (Table \ref{table:dmc-core}). All methods are trained for 500K steps with dense task-specific rewards. SAC+crop is the same as \secant's expert, from which the student distills for up to 30K steps without reward. SAC+IDM and PAD numbers are from \citet{hansen2020self}.

\para{Choice of student augmentations (Table \ref{table:dmc-all-augs}).}
We hypothesize that \secant needs multiple kinds of augmentations to resist a wide variety of distribution shifts at test time. Keeping the experts fixed, we study the effect of 6 different augmentations and their combinations on the student. In the challenging random video environments, Mixup and Cutmix tend to outperform other single operators. In most tasks, sampling from a mixture of augmentations generalizes better than solos, thus confirming our hypothesis. We adopt Combo1 for all \secant results in Table \ref{table:dmc-core}. %

\begin{table}[t!]
\vskip -0.1in
\caption{Ablation on imitation strategies. DAgger outperforms Expert-only and Student-only data collection in the second stage.
}
\label{table:dmc-exploration}
\vskip 0.1in
\centering
\scriptsize

\resizebox{\columnwidth}{!}{
\begin{tabular}{c|r|ccc}
\toprule
Setting & Task & DAgger & Expert & Student \\ \midrule 

\multirow{2}{*}{\makecell{Random \\ Color}} & Cheetah run & $\bestscore{582\pm64}$ & $519\pm73$ & $347\pm326$ \\
 & Walker walk & $\bestscore{856\pm31}$ & $818\pm41$ & $854\pm33$ \\
\midrule 

\multirow{2}{*}{\makecell{Random \\ Video}} & Cheetah run & $\bestscore{428\pm70}$ & $291\pm41$ & $264\pm241$ \\
 & Walker walk & $\bestscore{842\pm47}$ & $778\pm67$ & $822\pm55$ \\

\bottomrule
\end{tabular}
}

\end{table}

\begin{table}[t!]
\vskip -0.1in
\caption{Ablation on \secant-Parallel variant. It is advantageous to train expert and student sequentially rather than in parallel.
}
\label{table:dmc-parallel-variant}
\vskip 0.1in
\centering
\scriptsize

\resizebox{\columnwidth}{!}{
\begin{tabular}{c|r|cc}
\toprule
Setting & Task & \secant & \secant-Parallel \\ \midrule 

\multirow{4}{*}{\makecell{Random \\ Color}} & Cheetah run & $\bestscore{582\pm64\hphantom{0}}$ & $302\pm248$ \\
 & Ball in cup catch & $\bestscore{958\pm7\hphantom{00}}$ & $790\pm332$ \\
 & Cartpole swingup & $\bestscore{866\pm15\hphantom{0}}$ & $834\pm8\hphantom{00}$ \\
 & Walker walk & $\bestscore{856\pm31\hphantom{0}}$ & $768\pm22\hphantom{0}$ \\
\midrule 

\multirow{4}{*}{\makecell{Random \\ Video}} & Cheetah run & $\bestscore{428\pm70\hphantom{0}}$ & $276\pm216$ \\
 & Ball in cup catch & $\bestscore{903\pm49\hphantom{0}}$ & $676\pm280$ \\
 & Cartpole swingup & $752\pm38\hphantom{0}$ & $\bestscore{764\pm17\hphantom{0}}$  \\
 & Walker walk & $\bestscore{842\pm47\hphantom{0}}$ & $699\pm21\hphantom{0}$ \\

\bottomrule
\end{tabular}
}

\end{table}
\begin{table*}[t]
\vskip -0.07in
\caption{Robustness analysis in Robosuite: we measure the cycle consistency of observation embeddings across trajectories of the same task but different appearances. The higher the accuracy, the more robust the representation is against visual variations.
}
\label{table:robo-cycle}
\vskip 0.1in
\centering
\resizebox{0.97\textwidth}{!}{%
\begin{tabular}{c|r|ccccccc}

\toprule
Cycle & Tasks & \secant\, (Ours) & SAC & SAC+crop & DR & NetRand & SAC+IDM & PAD \\ \midrule 

\multirow{2}{*}{\makecell{2-way}} & Nut assembly & $\bestscore{77.3\pm7.6}$ \,(\bestpercent{+29.3}) & $24.0\pm6.0$ & $16.0\pm3.7$ & $25.3\pm11.9$ & $48.0\pm15.9$ & $29.3\pm13.0$ & $26.7\pm9.4$ \\
 & Two arm lifting & $\bestscore{72.0\pm9.9}$ \,(\bestpercent{+32.0}) & $20.0\pm0.0$ & $18.7\pm3.0$ & $24.0\pm10.1$ & $40.0\pm14.9$ & $18.7\pm3.0$ & $18.7\pm5.6$ \\
\midrule 

\multirow{2}{*}{\makecell{3-way}} & Nut assembly & $\bestscore{33.3\pm8.2}$ \,(\bestpercent{+17.3}) & $16.0\pm8.9$ & $16.0\pm3.7$ & $8.0\pm11.0$ & $9.3\pm10.1$ & $10.7\pm7.6$ & $10.7\pm7.6$ \\
 & Two arm lifting & $\bestscore{32.0\pm8.7}$ \,(\bestpercent{+20.0}) & $6.7\pm9.4$ & $2.7\pm3.7$ & $10.7\pm10.1$ & $6.7\pm6.7$ & $12.0\pm7.3$ & $12.0\pm7.3$ \\

\bottomrule
\end{tabular}
}
\vskip -0.1in
\end{table*}

\para{Single stage vs two-stage augmentation (Fig.~\ref{fig:dmc-weak-strong}).} 
The premise of \secant is that we cannot effectively apply strong augmentation (Combo1) in one stage to learn robust policies. We put this assumption to test. In Fig. \ref{fig:dmc-weak-strong}, ``S-only" and ``W-only" are single-stage policies trained with strong or weak augmentations. $X \rightarrow Y$ denotes two-stage training, e.g. S $\rightarrow$ W means a strongly-augmented expert is trained first, and then a weakly-augmented student imitates it.  
We highlight \textbf{4 key findings}: 
\textbf{(1)} single-stage RL trained with strong augmentation (S-only) underperforms in both training and test environments consistently, due to poor optimization. 
\textbf{(2)} The student is typically upper-bounded by the expert's performance, thus both S $\rightarrow$ W and S $\rightarrow$ S produce sub-optimal policies. 
\textbf{(3)} single-stage policy trained with random cropping (W-only) overfits on the training environment and generalizes poorly. Adding a weakly-augmented student (W $\rightarrow$ W) does not remedy the overfitting.
\textbf{(4)} The only effective strategy is a weakly-augmented expert followed by a strongly-augmented student (\narrowcolorbox{yellow}{W $\rightarrow$ S}), which is exactly \secant. It recovers the strong performance on the training environment, and bridges the generalization gap in unseen test environments. We include more extensive ablation results with different 2-stage augmentation strategies in Appendix \ref{appendix:aug-ablations}.

\para{Ablation on imitation strategies (Table \ref{table:dmc-exploration}).}
\secant uses DAgger (Sec. \ref{sec:preliminaries}) in the second stage, which rolls out the expert policy to collect initial trajectories, and then follows the student's policy. The alternatives are using expert or student policy alone to collect all trajectory data. The former approach lacks data diversity, while the latter slows down learning due to random actions in the beginning. Table \ref{table:dmc-exploration} validates the choice of DAgger for policy distillation.

\para{Ablation on the parallel-distillation variant (Table \ref{table:dmc-parallel-variant}).}
Can we train the expert and the student at the same time, rather than sequentially? We consider a variant of our method, called \secant-Parallel, that trains the expert alongside the student while keeping all other settings fixed. Similar to \secant, it also enjoys the nice property of disentangling robust representation learning from policy optimization. However, the student in \secant distills from a fully-trained and frozen expert, while the student in \secant-Parallel has to distill from a non-stationary expert, which leads to suboptimal performances. Table \ref{table:dmc-parallel-variant} demonstrates that it is more beneficial to adopt our proposed two-stage procedure, as \secant outperforms \secant-Parallel in a majority of tasks. We include more \secant-Parallel results on Robosuite in Appendix \ref{appendix:secant-parallel}.

\begin{figure}[t!] 
\small
\centering
\setlength\tabcolsep{1pt}
\begin{tabularx}{0.98\columnwidth}{ c | c c c }

\includegraphics[width=0.23\linewidth]{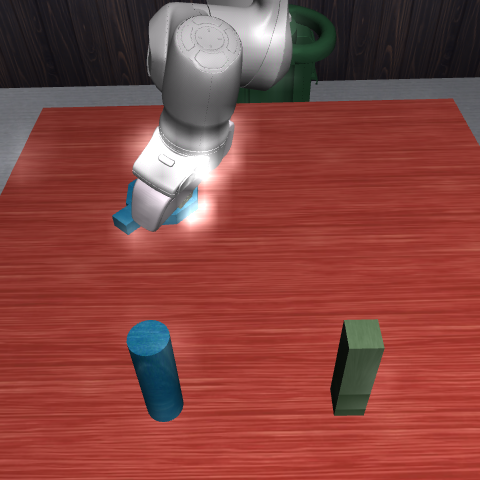} &
\includegraphics[width=0.23\linewidth]{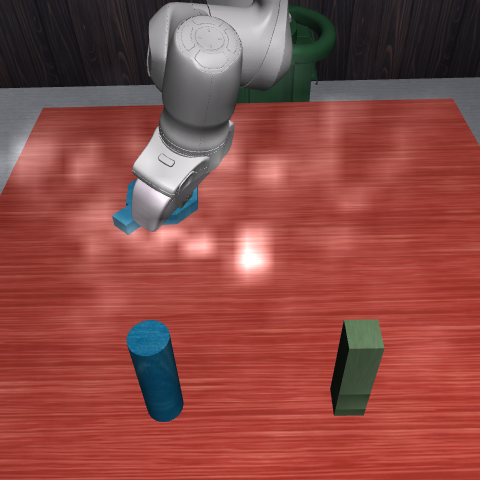} &
\includegraphics[width=0.23\linewidth]{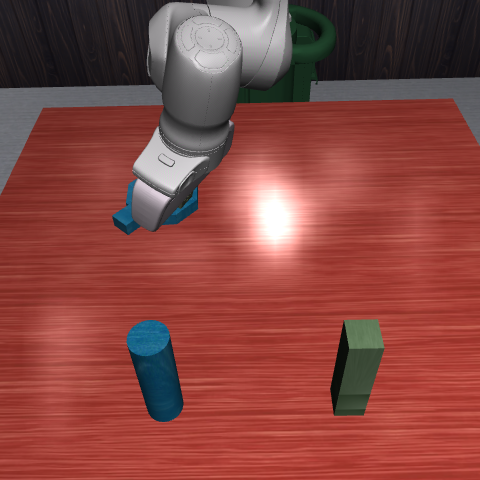} &
\includegraphics[width=0.23\linewidth]{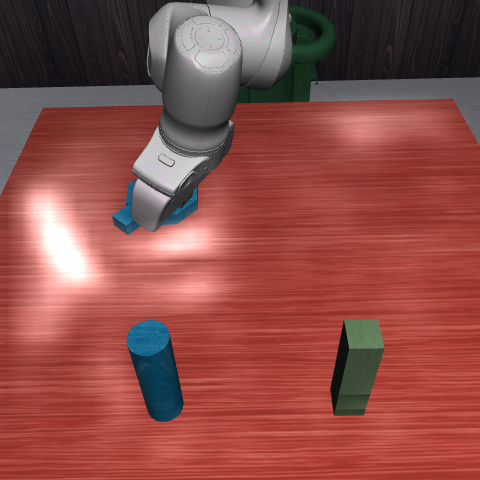} \\

\includegraphics[width=0.23\linewidth]{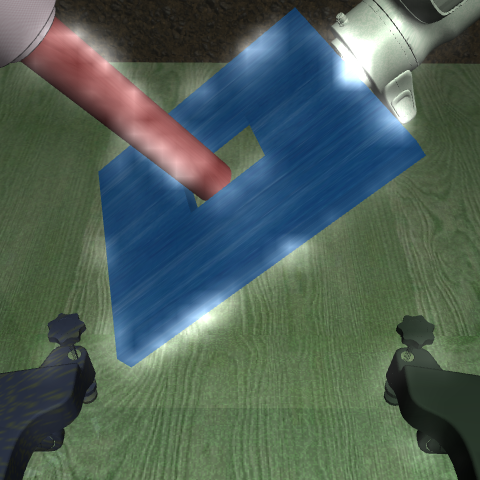}  &
\includegraphics[width=0.23\linewidth]{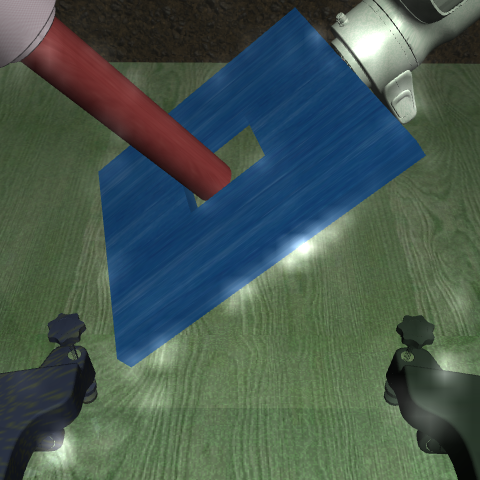} &
\includegraphics[width=0.23\linewidth]{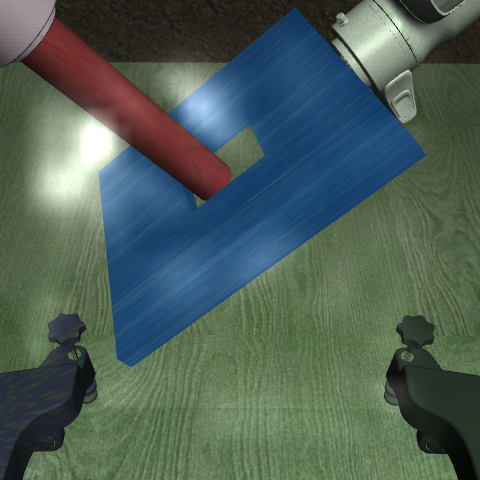} &
\includegraphics[width=0.23\linewidth]{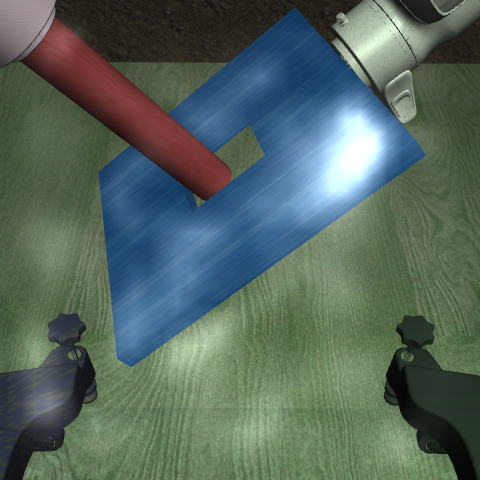} \\

\includegraphics[width=0.23\columnwidth]{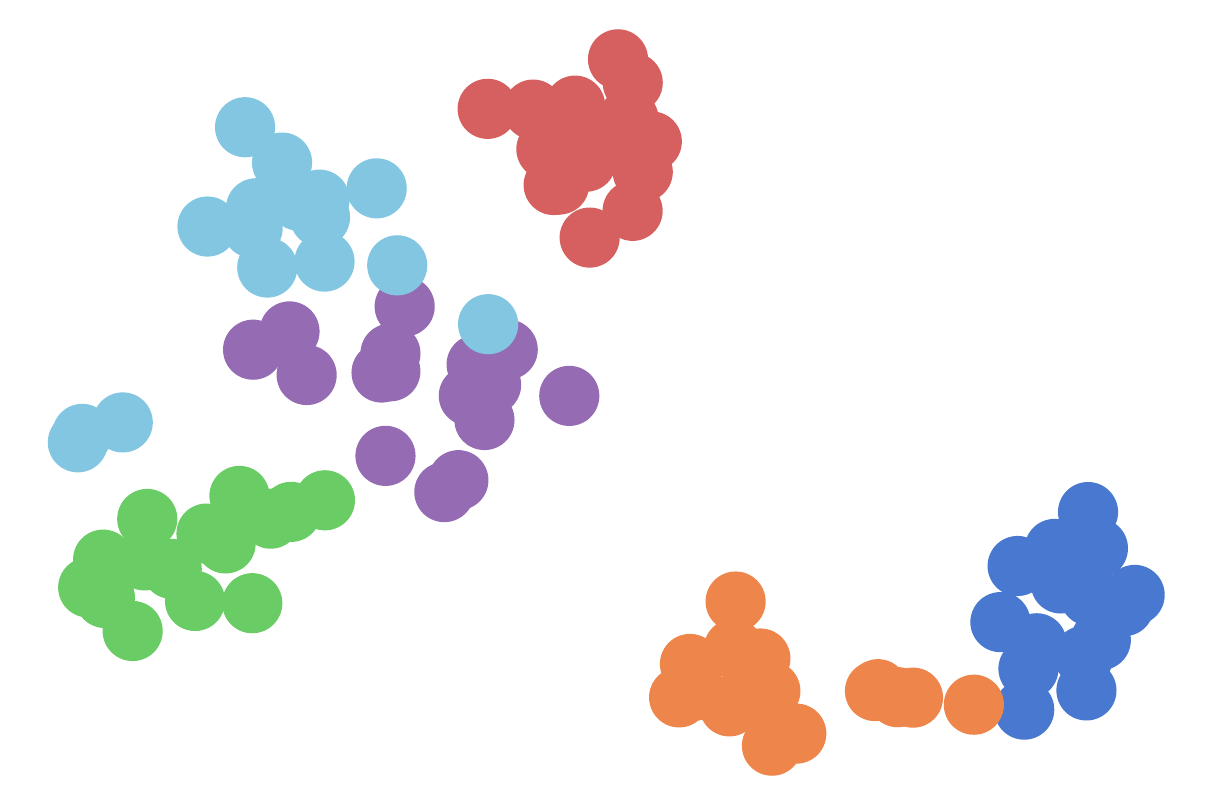}  &  
\includegraphics[width=0.23\columnwidth]{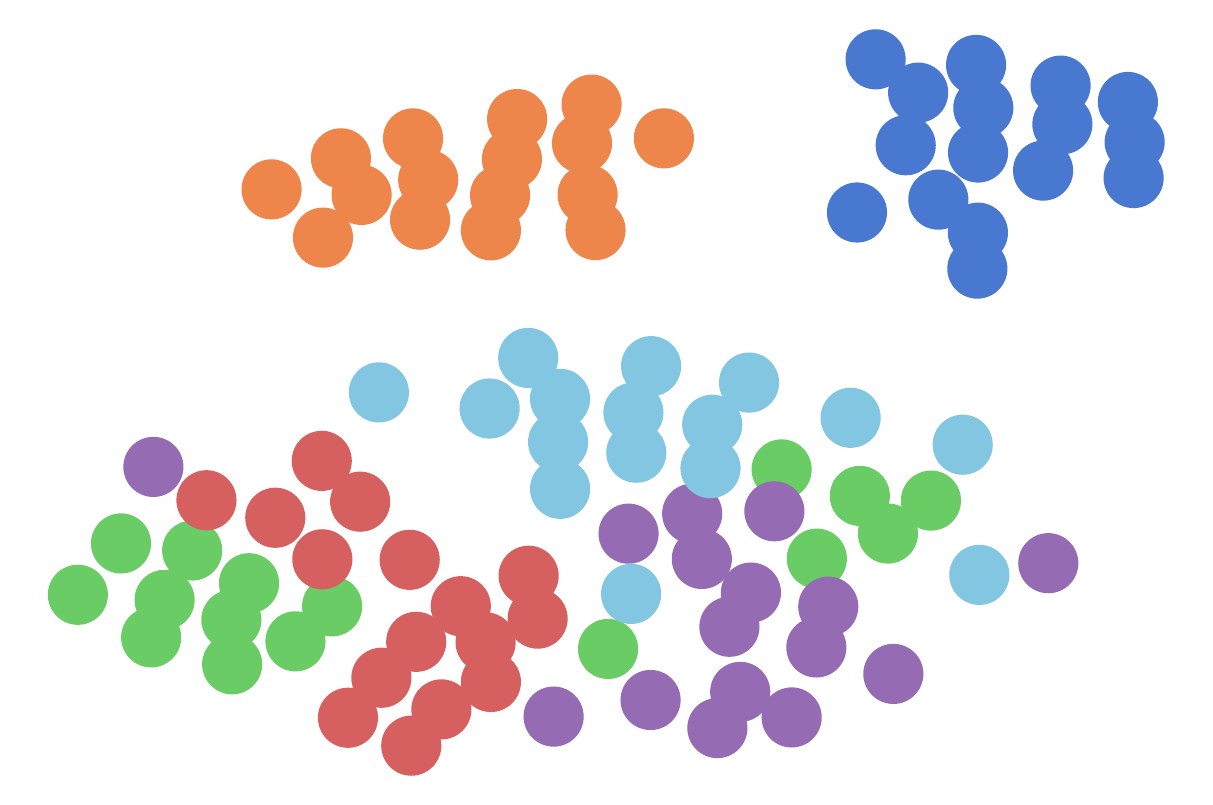} &
\includegraphics[width=0.23\columnwidth]{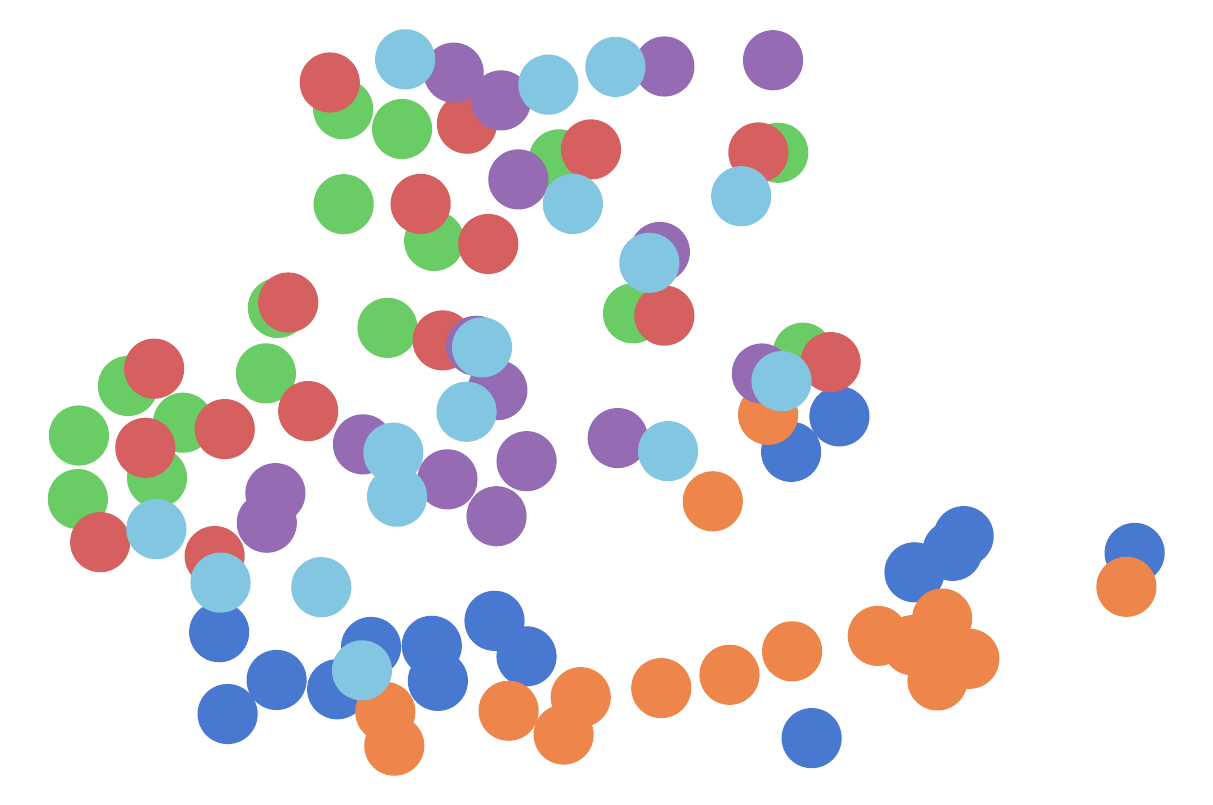} &
\includegraphics[width=0.23\columnwidth]{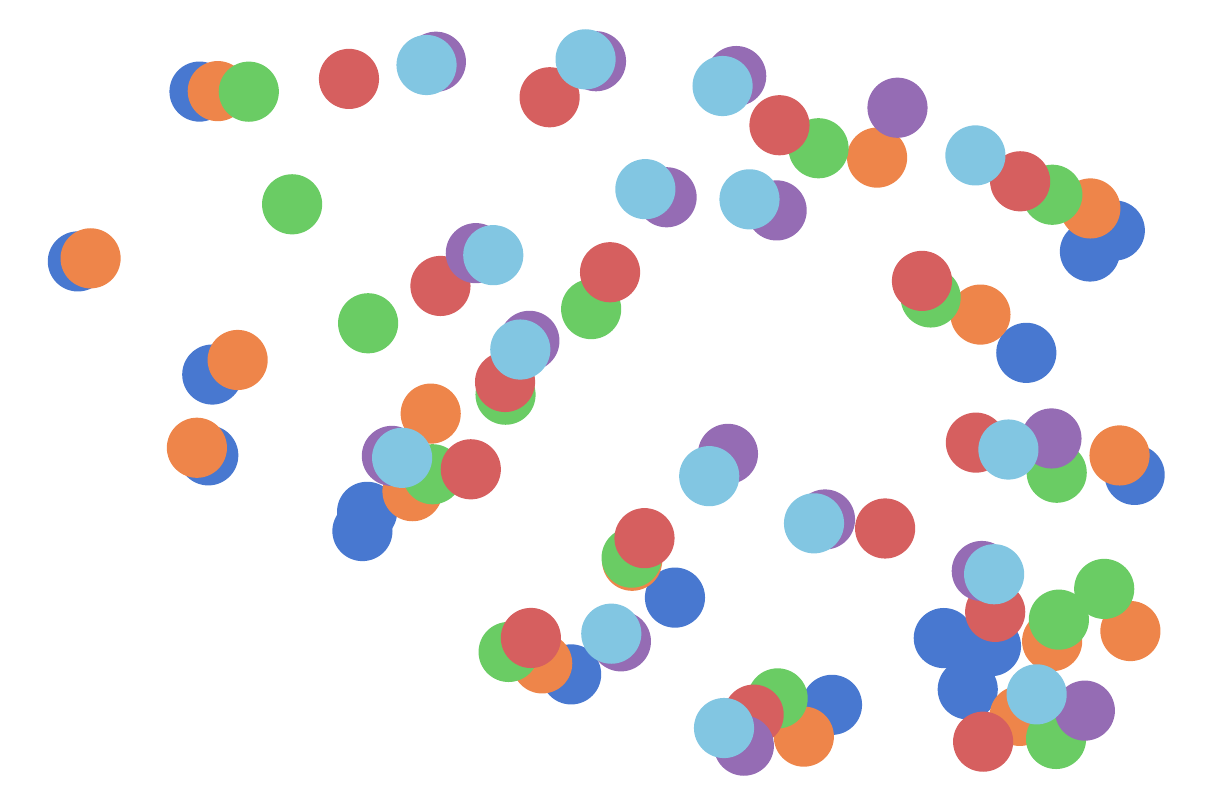} \\

\secantcaption & NetRand & PAD & DR 

\end{tabularx}
\caption{Row 1 and 2: saliency map of the learned policies in unseen tests. \secant attends to the components crucial to the task, while other agents often focus on irrelevant places. 
Row 3: t-SNE visualization of state embeddings. Our method correctly groups semantically similar states with different visual appearances.
}
\label{fig:robo-tsne-attn}
\end{figure}

\begin{table*}[t]
\caption{CARLA autonomous driving. The different weathers feature highly realistic raining, shadow, and sunlight changes. We report distance (m) travelled in a town without collision. \secant drives \textbf{+47.7\%} farther on average than other agents at test time.
}
\label{table:carla-core}
\vskip 0.1in
\centering
\resizebox{0.97\textwidth}{!}{%
\begin{tabular}{c|r|ccccccc}

\toprule

Setting & Weather & \secant\, (Ours) & SAC & SAC+crop & DR & NetRand & SAC+IDM & PAD \\ \midrule 

\multirow{1}{*}{\makecell{Training}} & Clear noon & $596\pm77\hphantom{0}$ \,\hphantom{(+00.0\%)}\hphantom{0} & $282\pm71\hphantom{0}$ & $\bestscore{684\pm114}$ & $486\pm141$ & $648\pm61\hphantom{0}$ & $582\pm96\hphantom{0}$ & $632\pm126$ \\
\midrule 

\multirow{5}{*}{\makecell{Test \\ Weathers}} & Wet sunset & $\bestscore{397\pm99\hphantom{0}}$ \,(\bestpercent{+\hphantom{0}39.8}) & $\hphantom{0}57\pm14\hphantom{0}$ & $\hphantom{0}26\pm18\hphantom{0}$ & $\hphantom{00}9\pm11\hphantom{0}$ & $284\pm84\hphantom{0}$ & $\hphantom{0}25\pm11\hphantom{0}$ & $\hphantom{0}36\pm12\hphantom{0}$ \\
 & Wet cloudy noon & $\bestscore{629\pm204}$ \,(\bestpercent{+\hphantom{00}5.7}) & $180\pm45\hphantom{0}$ & $283\pm85\hphantom{0}$ & $595\pm260$ & $557\pm107$ & $433\pm105$ & $515\pm52\hphantom{0}$ \\
 & Soft rain sunset & $\bestscore{435\pm66\hphantom{0}}$ \,(\bestpercent{+\hphantom{0}73.3}) & $\hphantom{0}55\pm28\hphantom{0}$ & $\hphantom{0}38\pm25\hphantom{0}$ & $\hphantom{0}25\pm41\hphantom{0}$ & $251\pm104$ & $\hphantom{0}36\pm32\hphantom{0}$ & $\hphantom{0}41\pm37\hphantom{0}$ \\
 & Mid rain sunset & $\bestscore{470\pm202}$ \,(\bestpercent{+101.7}) & $\hphantom{0}50\pm8\hphantom{00}$ & $\hphantom{0}37\pm16\hphantom{0}$ & $\hphantom{0}24\pm24\hphantom{0}$ & $233\pm117$ & $\hphantom{0}42\pm23\hphantom{0}$ & $\hphantom{0}32\pm21\hphantom{0}$ \\
 & Hard rain noon & $\bestscore{541\pm96\hphantom{0}}$ \,(\bestpercent{+\hphantom{0}18.1}) & $237\pm85\hphantom{0}$ & $235\pm129$ & $341\pm96\hphantom{0}$ & $458\pm72\hphantom{0}$ & $156\pm194$ & $308\pm141$ \\
 
\bottomrule
\end{tabular}
}
\end{table*}

\newcommand{\addIGFig}[0]{\includegraphics[width=\linewidth]{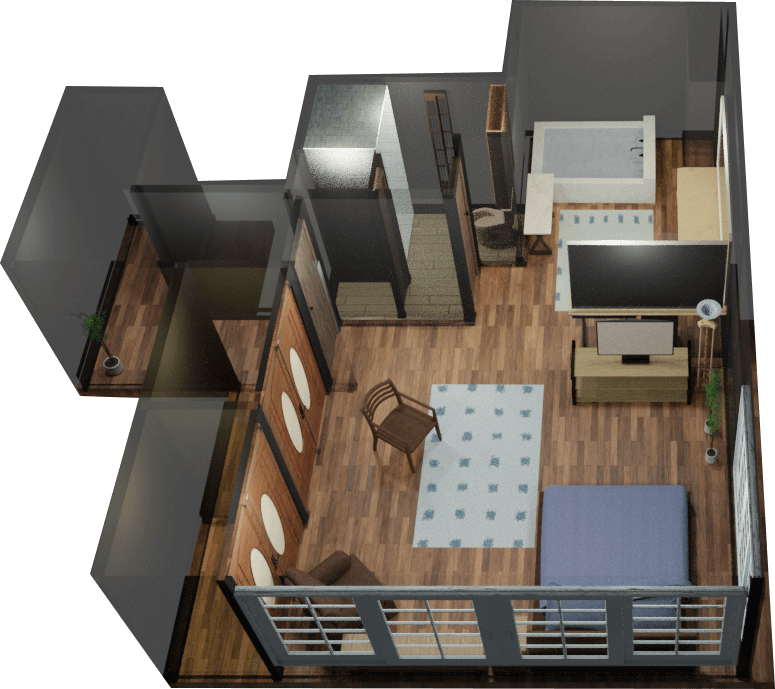}}

\begin{table*}[t]
\caption{iGibson object navigation. The goal is to find and navigate to a ceiling lamp in unseen rooms with novel decoration, furniture, and layouts (sample floor plan below). In testing, \secant has \textbf{+15.8\%} higher success rate (absolute percentage) than competing methods.
}
\label{table:ig-core}
\vskip 0.1in
\centering
\resizebox{\textwidth}{!}{%
\begin{tabular}{C{0.13\textwidth}c|cccccccc}

\toprule
\vspace{-0.13in}

\multirow{4}{*}{\makecell{\addIGFig{}}} & Setting & \secant\, (Ours) & SAC & SAC+crop & DR & NetRand & SAC+IDM & PAD \\ \cmidrule(lr){2-9}

& \multirow{1}{*}{\makecell{Training}} & $64.0\pm3.7\hphantom{0}$\,\,\hphantom{(+00.0\%)} & $\bestscore{68.7\pm2.5}$ & $51.0\pm12.0$ & $49.6\pm12.7$ & $56.4\pm3.8$ & $54.2\pm8.8$ & $59.0\pm13.4$ \\
\cmidrule(lr){2-9}  

& \multirow{1}{*}{\makecell{Test: Easy}} & $\bestscore{56.8\pm17.2}$ \,(\bestpercent{+17.6}) & $13.8\pm7.5$ & $12.9\pm7.1$ & $17.6\pm13.2$ & $39.2\pm11.7$ & $25.9\pm12.4$ & $30.9\pm12.4$ \\

& \multirow{1}{*}{\makecell{Test: Hard}} & $\bestscore{47.7\pm11.3}$ \,(\bestpercent{+13.9}) & $9.3\pm7.6$ & $7.9\pm5.3$ & $15.2\pm15.3$ & $33.8\pm11.8$ & $12.7\pm8.3$ & $26.1\pm23.0$ \\

\bottomrule

\end{tabular}
}
\end{table*}

\subsection{Robosuite: Robotic Manipulation}

Robosuite \cite{zhu2020robosuite} is a modular simulator for robotic research. We benchmark \secant and prior methods on 4 challenging single-arm and bimanual manipulation tasks. We use the Franka Panda robot model with operational space control, and train with task-specific dense reward. All agents receive a 168 $\!\times\!$ 168 egocentric RGB view as input (example in Table \ref{table:robo-core}, high-res version in Appendix \ref{appendix:robosuite}). The positions of moveable objects are randomized in each episode.  
\textbf{Door opening}: a robot arm must turn the handle and open the door in front of it. 
\textbf{Nut assembly}: two colored pegs (square and round) are mounted on the tabletop. The robot must fit the round nut onto the round peg. 
\textbf{Two-arm lifting}: two arms on opposite ends must each grab a handle of a large pot and lift it up above certain height.
\textbf{Peg-in-hole}: one arm holds a board with a square hole in the center, and the other holds a long peg. Two arms must coordinate to insert the peg into the hole. 

All agents are trained with clean background and objects, and evaluated on 3 progressively harder sets of environments (Table \ref{table:robo-core}). We design 10 variations for each task and difficulty level, and report the mean reward over 100 evaluation episodes (10 per variation). Reward below 100 indicates a failure to solve the task.  
\secant gains an average of \textbf{+287.5\%} more reward in \textit{easy} set, \textbf{+374.3\%} in \textit{hard} set, and \textbf{+351.6\%} in \textit{extreme} set over the best prior method.
The \textit{hard} and \textit{extreme} settings are particularly challenging because the objects of interest are difficult to discern from the noisy background. For nut assembly and two-arm lifting, \secant is the only agent able to obtain non-trivial partial rewards in \textit{hard} and \textit{extreme} modes consistently.

\para{Embedding robustness analysis.} 
To verify that our method develops high-quality representation, we measure the cycle consistency metric proposed in \citet{aytar2018playing}.  
First, given two trajectories $U$ and $V$, observation $u_i\in U$ locates its nearest neighbor in $V$:  $v_j=\arg\min_{v\in V} \left\lVert \phi(u_i) - \phi(v)\right\rVert^2$, where $\phi(\cdot)$ denotes the 50-D embedding from the visual encoder of the learned policies. 
Then in reverse, $v_j$ finds its nearest neighbor from $U$: $u_k=\arg\min_{u\in U} \left\lVert \phi(u) - \phi(v_j)\right\rVert^2$.
$u_i$ is cycle consistent if and only if $|i - k| \leq 1$, i.e. it returns to the original position. High cycle consistency indicates that the two trajectories are accurately aligned in the embedding space, despite their visual appearance shifts. We also evaluate 3-way cycle consistency that involves a third trajectory $W$, and measure whether $u_i$ can return to itself along \textit{both} $U \!\rightarrow\! V \!\rightarrow\! W \!\rightarrow\! U$ and $U \!\rightarrow\! W \!\rightarrow\! V \!\rightarrow\! U$. In Table \ref{table:robo-cycle}, we sample 15 observations from each trajectory, and report the mean cycle consistency over 5 trials.  
In Fig. \ref{fig:robo-tsne-attn}, we also visualize the state embeddings of door-opening task with t-SNE \cite{maaten2008tsne}. 
Both quantitative and qualitative analyses show that \secant significantly improves the robustness of visual representation over the baselines.

\para{Saliency visualization.} To better understand how the agents execute their policies, we compute saliency maps as described in \citet{greydanus2018visualizingrlattention}. We add Gaussian perturbation to the observation image at every $5 \!\times\! 5$ pixel patch, and visualize the saliency patterns in Fig. \ref{fig:robo-tsne-attn}. \secant is able to focus on the most task-relevant objects, even with novel textures it has not encountered during training.

\subsection{CARLA: Autonomous Driving}

To further validate \secant's generalization ability on natural variations, we construct a realistic driving scenario with visual observations in the CARLA simulator \cite{dosovitskiy2017carla}. 
The goal is to drive as far as possible along a figure-8 highway (CARLA Town 4) in 1000 time steps without colliding into 60 moving pedestrians or vehicles.
Our reward function is similar to \citet{amyzhang2020invariant}, which rewards progression, penalizes collisions, and discourages abrupt steering. 
The RGB observation is a 300-degree panorama of $84 \!\times\! 420$ pixels, formed by concatenating 5 cameras on the vehicle's roof with 60-degree view each. The output action is a 2D vector of thrust (brake is negative thrust) and steering. 

The agents are trained at ``clear noon", and evaluated on a variety of dynamic weather and lighting conditions at noon and sunset (Fig. \ref{fig:benchmark}). For instance, the \textit{wet} weathers feature roads with highly reflective spots. Averaged over 10 episodes per weather and 5 training runs, \secant is able to drive \textbf{+47.7\%} farther than prior SOTAs in tests.

\para{Inference speed}. The latency between observing and acting is critical for safe autonomous driving. Unlike \secant, PAD requires extra inference-time gradient computation. We benchmark both methods on actual hardware. Averaged over 1000 inference steps, \secant is \textbf{65 $\!\times$} faster than PAD on Intel Xeon Gold 5220 (2.2 GHz) CPU, and \textbf{42 $\!\times$} faster on Nvidia RTX 2080Ti GPU.

\subsection{iGibson: Indoor Object Navigation}

iGibson \cite{xia2020igibson,shen2020igibsonicra} is an interactive simulator with highly realistic 3D rooms and furniture (Fig. \ref{fig:benchmark}).
The goal is to navigate to a lamp as closely as possible. The reward function incentivizes the agent to maximize the proportion of pixels that the lamp occupies in view, and success is achieved when this proportion exceeds 5\% over 10 consecutive steps. 
Our benchmark features 1 training room and 20 test rooms, which include distinct furniture, layout, and interior design from training. 
The lamp is gray in training, but has much richer textures in testing. We construct 2 difficulty levels with 10 rooms each, depending on the extent of visual shift. The agent is randomly spawned in a room with \textit{only} RGB observation ($168\!\times\!168$), and outputs a 2D vector of linear and angular velocities. 

We evaluate on each test room for 20 episodes and report success rates in Table \ref{table:ig-core}. SAC without augmentation is better than SAC+crop because the lamp can be cropped out accidentally, which interferes with the reward function. Therefore we use plain SAC as the expert for \secant. We consider this an edge case, since random cropping is otherwise broadly applicable. \secant achieves \textbf{+15.8\%} higher success rate than prior methods in unseen rooms.

\section{Conclusion}

Zero-shot generalization in visual RL has been a long-standing challenge. We introduce \secant, a novel technique that addresses policy optimization and robust representation learning \textit{sequentially}. We demonstrate that \secant significantly outperforms prior SOTA in 4 challenging domains with realistic test-time variations.
We also systematically study different augmentation recipes, strategies, and distillation approaches. Compared to prior methods, we find that \secant develops more robust visual representations and better task-specific saliency maps.

\section*{Acknowledgements}

We are extremely grateful to Chris Choy, Jean Kossaifi, Shikun Liu, Zhiyuan ``Jerry" Lin, Josiah Wong, Huaizu Jiang, Guanya Shi, Jacob Austin, Ismail Elezi, Ajay Mandlekar, Fei Xia, Agrim Gupta, Shyamal Buch, and many other colleagues for their helpful feedback and insightful discussions.

\bibliography{ms}
\bibliographystyle{icml2021}

\newpage
\appendix
\section{Algorithm Details}
\label{appendix:algorithm-hypers}

In this section, we provide a detailed account of our algorithm implementation and hyperparameters.  

We implement Soft Actor Critic \cite{haarnoja2018soft,haarnoja2018soft2} for all environments in PyTorch v1.7 \cite{paszke2019pytorch} with GPU acceleration for visual observations. We follow the random cropping scheme in \citet{kostrikov2020image}, which first applies a reflection padding to expand the image, and then crop back to the original size. Some of the augmentation operators are implemented with the Kornia library \cite{riba2020kornia}. 

All training hyperparameters are listed in Table \ref{table:hyper}. We perform a small grid search for the learning rates, and tune them only on the training environment. 
All agents are optimized by Adam \cite{Kingma2015AdamAM} with the default settings ($\beta_1=0.9, \beta_2=0.999, \epsilon=10^{-8}$) in PyTorch.

\begin{table*}
\caption{\secant hyperparameters for all environments. }
\label{table:hyper}
\vskip 0.1in
\centering
\resizebox{0.8\textwidth}{!}{%
\begin{tabular}{@{}lc|c|c|c@{}}
\toprule
Hyperparameter & DMControl & Robosuite & CARLA & iGibson\\ 
\midrule
Input dimension & $9\times84\times84$ & $9\times168\times168$ & $9\times84\times420$ & $9\times168\times168$\\
Stacked frames & 3 & 3 & 3 & 3 \\
Discount factor $\gamma$ & 0.99 & 0.99 & 0.99 & 0.99 \\
Episode length & 1000 & 500 & 1000 & 500 \\
Number of training steps & 500K & 800K & 500K & 800K\\ 
SAC replay buffer size & 100K & 100K & 100K & 50K\\
SAC batch size & 512 & 512 & 1024 & 128 \\
Optimizer & Adam & Adam & Adam & Adam \\
Actor learning rate & 5e-4 (Walker-walk) & 1e-4 (Peg-in-hole) & 1e-3 & 5e-4\\
& 1e-3 (otherwise) & 1e-3 (otherwise) & & \\ 
Critic learning rate & 5e-4 (Walker-walk) & 1e-4 & 1e-3 & 1e-4 \\
& 1e-3 (otherwise) & & & \\ 
$\log\alpha$ learning rate & 5e-4 (Walker-walk) & 1e-4 (Peg-in-hole) & 1e-3 & 5e-4\\
& 1e-3 (otherwise) & 1e-3 (otherwise) & &  \\ 
Critic target update frequency & 2 & 4 & 2 & 4\\
Random cropping padding & 4 & 8 & (4, 12) & 8 \\
Encoder conv layers & 4 & 4 & 4 & 4 \\
Encoder conv strides & [2, 1, 1, 2] & [2, 1, 1, 2] & [2, (1, 2), (1, 2), 2]  & [2, 2, 1, 1]\\
Encoder conv channels & 32 & 32 & [64, 64, 64, 32] & 32 \\
Encoder feature dim & 50 & 50 & 64 & 50\\
Actor head MLP layers & 3 & 3 & 3 & 3 \\
Actor head MLP hidden dim & 1024 & 1024 & 1024 & 1024 \\
\secant student augmentation & Combo1 & Combo2 & Combo1 & Combo2 \\
\secant learning rate & 1e-3 & 1e-3 & 1e-3 & 1e-3 \\
\secant replay buffer size & 10K & 20K & 10K & 20K\\
\secant batch size & 512 & 512 & 1024 & 512\\
\bottomrule
\end{tabular}
}
\end{table*}

\subsection{Inference Latency}

At inference time, latency between observing and acting is crucial for real-world deployment. Unlike \secant that only performs a single forward pass, PAD \cite{hansen2020self} requires expensive test-time gradient computation. In addition, PAD needs a deeper ConvNet as encoder and extra test-time image augmentations for the auxiliary self-supervised mechanism to work well \cite{hansen2020self}. These add even more overhead during inference. 

\begin{figure}[t!] \small
\centering
\setlength\tabcolsep{1.5pt}
\begin{tabularx}{\linewidth}{ c c }

CPU & GPU \\ 

\includegraphics[width=0.49\linewidth]{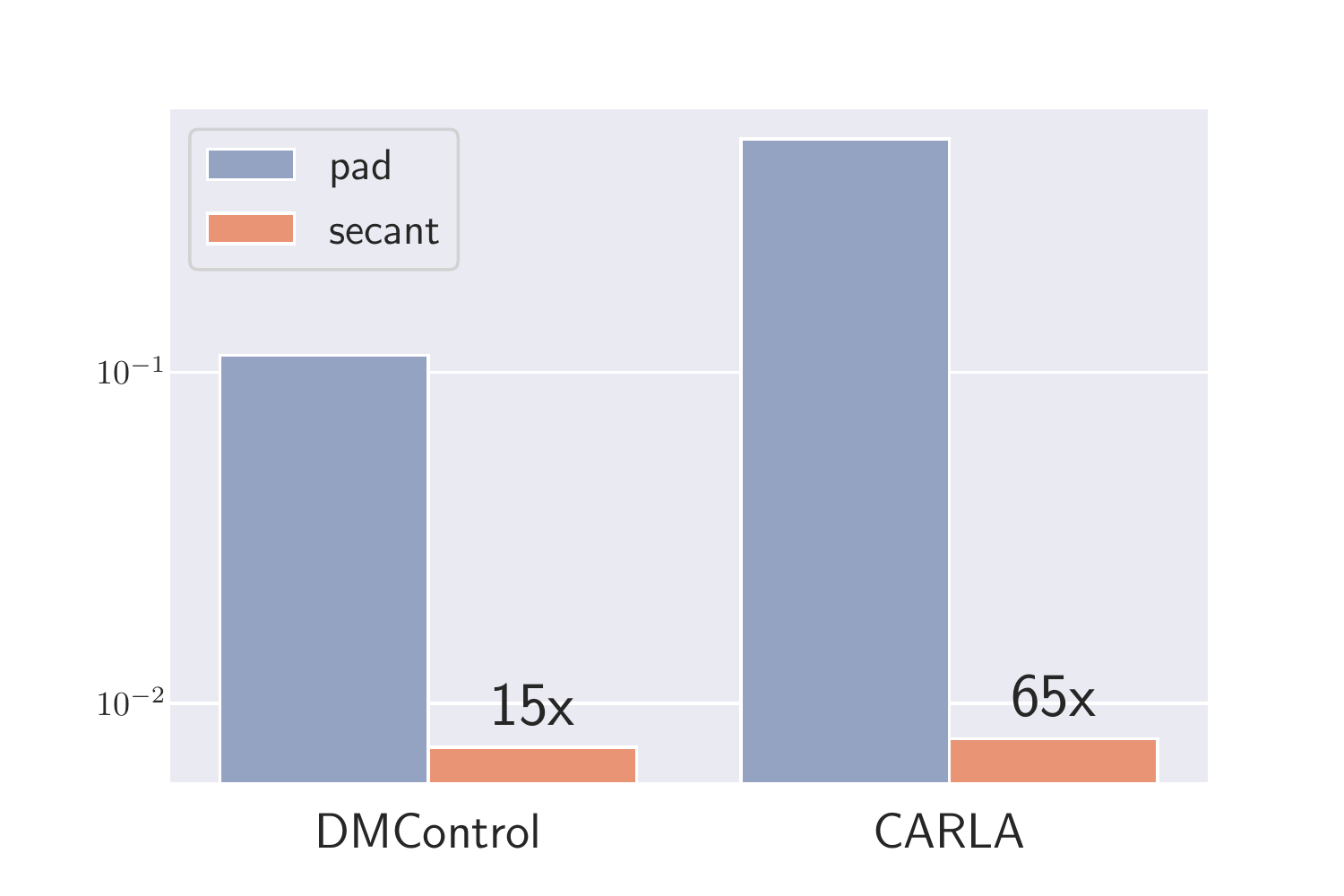}  &  
\includegraphics[width=0.49\linewidth]{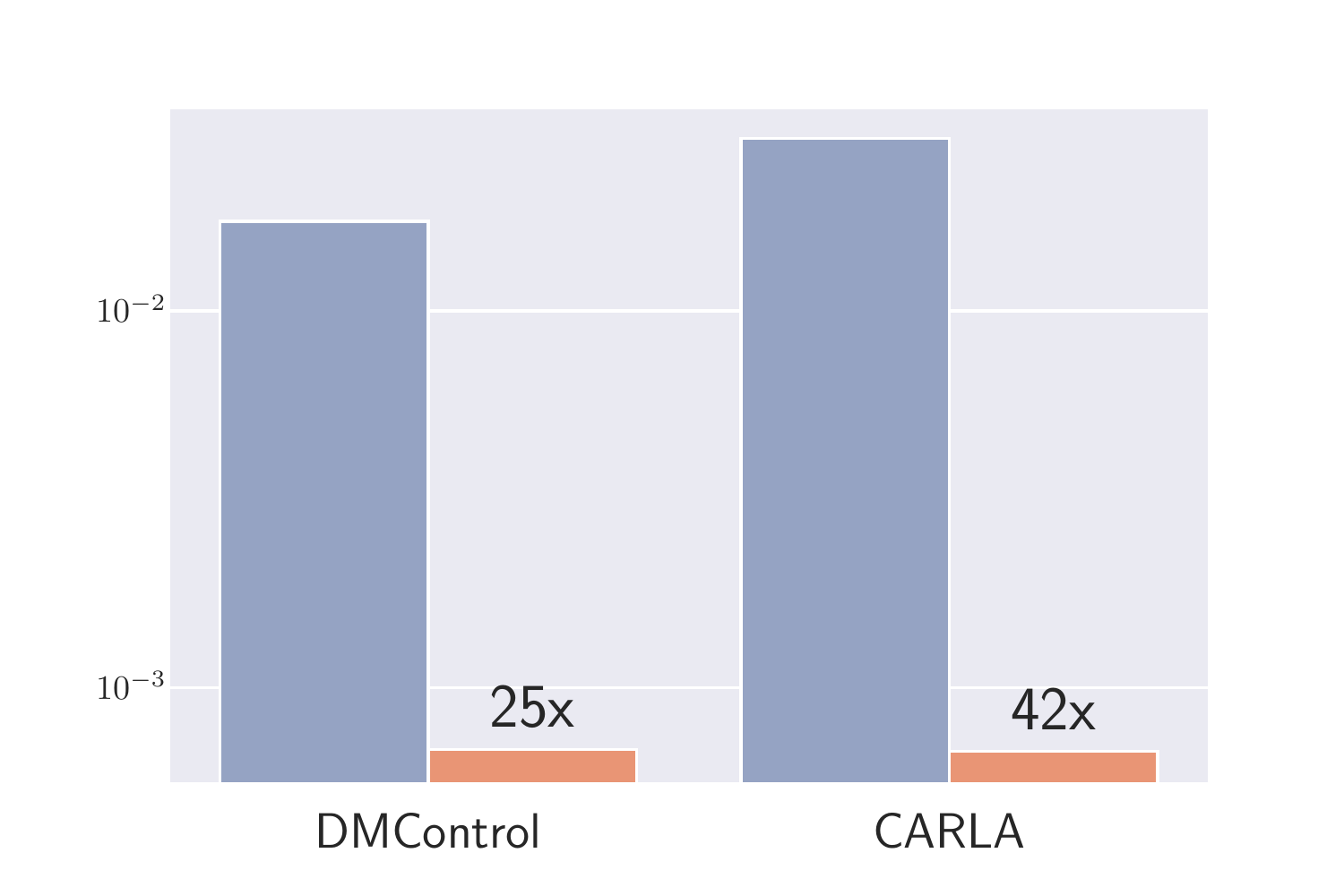}   
\end{tabularx}
\caption{\secant vs PAD inference latency. Y-axis denotes average seconds per action (log-scale). \secant improves inference speed by an order of magnitude compared to PAD. 
}
\label{fig:speed}
\end{figure}

We expand on Section~5.3 in the main paper, and benchmark both \secant and PAD on actual hardware for DMControl and CARLA (Fig.~\ref{fig:speed}). The CPU model is Intel Xeon Gold 5220 (2.2 GHz) CPU, and the GPU model is Nvidia RTX 2080Ti. The latency is averaged over 1000 inference steps and excludes the simulation time. We show that \secant improves inference speed by an order of magnitude in both environments.

\section{Environment Details}

\subsection{Deepmind Control Suite}

We follow the environment settings introduced in \citet{hansen2020self}. We use 8 tasks that support randomized colors and 7 tasks that support distracting video background (\texttt{Reacher-easy} does not support inserting videos). We use the same action repeat settings as \citet{hansen2020self}: 2 for \texttt{Finger-spin}, 8 for \texttt{Cartpole-swingup} and \texttt{Cartpole-balance}, and 4 for the rest. Please refer to \citet{deepmindcontrolsuite2018} and \citet{hansen2020self} for more details about DMControl tasks.

\subsection{Robosuite}
\label{appendix:robosuite}

\begin{figure*}
    \centering
    \includegraphics[width=0.93\textwidth]{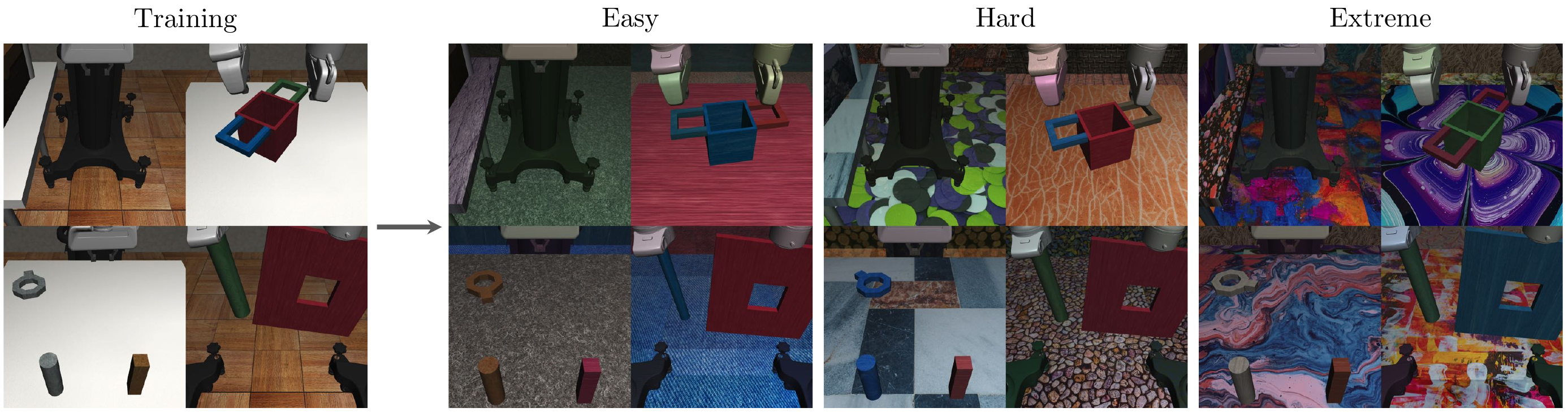}
    \caption{Sample Robosuite environments. Tasks in clockwise order: Door opening, Two-arm lifting, Peg-in-hole, and Nut assembly.}
    \label{fig:robo-env}
\end{figure*}

We use the Franka Panda robot model with operational space control for 4 Robosuite tasks. 
The action dimensions for door opening, nut assembly, peg-in-hole, and two-arm lifting are 7, 7, 12 and 14, respectively. We use continuous action space. The control frequency is set to 20Hz, which means that the robot receives 20 control signals during every simulation second.
We provide brief descriptions of each task and their associated reward functions below. All environments add an extra positive reward upon task completion, in addition to the dense reward shaping. Example observations are shown in Fig.~\ref{fig:robo-env}. Please refer to \citet{zhu2020robosuite} for more details.  

\para{Door opening.}
A robot arm must learn to turn the handle and open the door in front of it. The reward is shaped by the distance between the door handle and the robot arm, and the rotation angle of the door handle. 

\para{Nut assembly.}
Two colored pegs (one square and one round) are mounted on the tabletop. The robot must fit the round nut onto the round peg. At first, the robot receives a reaching reward inversely proportional to the distance between the gripper and the nut, and a binary reward once it grasps the nut successfully. After grasping, it obtains a lifting reward proportional to the height of the nut, and a hovering reward inversely proportional to the distance between the nut and the round peg. 

\para{Peg-In-Hole.}
One arm holds a board with a square hole in the center, and the other holds a long peg. The two arms must coordinate to insert the peg into the hole. The reward is shaped by the distance between two arms, along with the orientation and distance between the peg and the hole. 

\para{Two-arm lifting.} 
A large pot with two handles is placed on the table. Two arms on opposite ends must each grab a handle and lift the pot together above certain height, while keeping it level. At first, the agent obtains a reaching reward inversely proportional to the distance between each arm and its respective pot handle, and a binary reward if each gripper is grasping the correct handle. After grasping, the agent receives a lifting reward proportional to the pot’s height above the table and capped at a certain threshold. 

No agent can solve nut assembly and two-arm lifting completely. However, \secant is able to obtain partial rewards by grasping the nut or the pot handles successfully in the unseen test environments, while the prior SOTA methods struggle. There is still room to improve on this challenging benchmark. 

\begin{table*}[t]
\caption{iGibson object navigation: detailed breakdown of \textit{Easy} and \textit{Hard} settings.
}
\label{table:ig-breakdown}
\vskip 0.1in
\centering
\resizebox{0.95\textwidth}{!}{%
\begin{tabular}{C{0.13\textwidth}|r|ccccccc}

\toprule

Setting & Room & \secant (Ours) & SAC & SAC+crop & DR & NetRand & SAC+IDM & PAD \\ \midrule 
\multirow{10}{*}{\specialcell{Easy \\ Rooms}} & Beechwood & $\bestscore{63.0\pm17.5}$ & $17.0\pm10.4$ & $16.0\pm8.2\hphantom{0}$ & $21.0\pm10.2$ & $47.0\pm7.6\hphantom{0}$ & $28.0\pm8.4\hphantom{0}$ & $33.0\pm13.0$ \\
 & Ihlen & $\bestscore{58.0\pm19.2}$ & $\hphantom{0}9.0\pm4.2\hphantom{0}$ & $11.0\pm8.2\hphantom{0}$ & $\hphantom{0}6.0\pm5.5\hphantom{0}$ & $42.0\pm9.1\hphantom{0}$ & $24.0\pm10.8$ & $25.0\pm7.9\hphantom{0}$ \\
 & Merom & $\bestscore{65.0\pm14.6}$ & $20.0\pm7.1\hphantom{0}$ & $15.0\pm6.1\hphantom{0}$ & $20.0\pm12.7$ & $38.0\pm9.7\hphantom{0}$ & $24.0\pm6.5\hphantom{0}$ & $26.0\pm6.5\hphantom{0}$ \\
 & Wainscott & $\bestscore{64.0\pm8.2\hphantom{0}}$ & $18.0\pm2.7\hphantom{0}$ & $15.0\pm7.1\hphantom{0}$ & $41.0\pm15.2$ & $49.0\pm14.7$ & $35.0\pm10.6$ & $29.0\pm12.4$ \\
 & Benevolence-0 & $\bestscore{67.0\pm15.7}$ & $15.0\pm3.5\hphantom{0}$ & $16.0\pm9.6\hphantom{0}$ & $17.0\pm8.4\hphantom{0}$ & $49.0\pm8.2\hphantom{0}$ & $39.0\pm11.4$ & $51.0\pm8.2\hphantom{0}$ \\
 & Benevolence-1 & $\bestscore{48.0\pm10.4}$ & $\hphantom{0}6.0\pm6.5\hphantom{0}$ & $13.0\pm5.7\hphantom{0}$ & $\hphantom{0}6.0\pm6.5\hphantom{0}$ & $34.0\pm8.2\hphantom{0}$ & $18.0\pm10.4$ & $21.0\pm8.2\hphantom{0}$ \\
 & Benevolence-2 & $\bestscore{59.0\pm22.2}$ & $14.0\pm8.2\hphantom{0}$ & $\hphantom{0}8.0\pm6.7\hphantom{0}$ & $21.0\pm11.9$ & $38.0\pm9.7\hphantom{0}$ & $31.0\pm17.1$ & $42.0\pm11.5$ \\
 & Pomaria-1 & $\bestscore{47.0\pm24.6}$ & $13.0\pm9.1\hphantom{0}$ & $10.0\pm7.1\hphantom{0}$ & $15.0\pm7.9\hphantom{0}$ & $27.0\pm10.4$ & $22.0\pm16.0$ & $34.0\pm7.4\hphantom{0}$ \\
 & Pomaria-2 & $\bestscore{58.0\pm14.4}$ & $16.0\pm6.5\hphantom{0}$ & $15.0\pm7.9\hphantom{0}$ & $14.0\pm10.2$ & $32.0\pm9.1\hphantom{0}$ & $22.0\pm9.1\hphantom{0}$ & $27.0\pm4.5\hphantom{0}$ \\
 & Rs & $\bestscore{39.0\pm8.2\hphantom{0}}$ & $10.0\pm6.1\hphantom{0}$ & $10.0\pm5.0\hphantom{0}$ & $15.0\pm9.4\hphantom{0}$ & $36.0\pm13.4$ & $16.0\pm8.2\hphantom{0}$ & $21.0\pm9.6\hphantom{0}$ \\
\midrule 
\multirow{10}{*}{\specialcell{Hard \\ Rooms}} & Beechwood & $\bestscore{61.0\pm12.4}$ & $\hphantom{0}7.0\pm5.7\hphantom{0}$ & $\hphantom{0}5.0\pm5.0\hphantom{0}$ & $10.0\pm3.5\hphantom{0}$ & $31.0\pm9.6\hphantom{0}$ & $\hphantom{0}7.0\pm2.7\hphantom{0}$ & $16.0\pm8.2\hphantom{0}$ \\
 & Ihlen & $\bestscore{37.0\pm14.0}$ & $\hphantom{0}4.0\pm4.2\hphantom{0}$ & $10.0\pm5.0\hphantom{0}$ & $\hphantom{0}6.0\pm5.5\hphantom{0}$ & $30.0\pm5.0\hphantom{0}$ & $\hphantom{0}9.0\pm8.2\hphantom{0}$ & $26.0\pm9.6\hphantom{0}$ \\
 & Merom & $\bestscore{45.0\pm5.0\hphantom{0}}$ & $\hphantom{0}4.0\pm4.2\hphantom{0}$ & $\hphantom{0}4.0\pm4.2\hphantom{0}$ & $10.0\pm11.7$ & $32.0\pm10.4$ & $11.0\pm4.2\hphantom{0}$ & $25.0\pm9.4\hphantom{0}$ \\
 & Wainscott & $\bestscore{46.0\pm9.6\hphantom{0}}$ & $10.0\pm7.1\hphantom{0}$ & $10.0\pm3.5\hphantom{0}$ & $10.0\pm5.0\hphantom{0}$ & $23.0\pm4.5\hphantom{0}$ & $12.0\pm5.7\hphantom{0}$ & $\hphantom{0}9.0\pm6.5\hphantom{0}$ \\
 & Benevolence-0 & $\bestscore{56.0\pm7.4\hphantom{0}}$ & $\hphantom{0}7.0\pm2.7\hphantom{0}$ & $\hphantom{0}6.0\pm2.2\hphantom{0}$ & $\hphantom{0}9.0\pm6.5\hphantom{0}$ & $36.0\pm8.2\hphantom{0}$ & $11.0\pm2.2\hphantom{0}$ & $11.0\pm8.9\hphantom{0}$ \\
 & Benevolence-1 & $44.0\pm4.2\hphantom{0}$ & $12.0\pm9.7\hphantom{0}$ & $\hphantom{0}8.0\pm5.7\hphantom{0}$ & $\bestscore{49.0\pm14.3}$ & $36.0\pm8.9\hphantom{0}$ & $13.0\pm10.4$ & $15.0\pm12.7$ \\
 & Benevolence-2 & $\bestscore{49.0\pm15.6}$ & $21.0\pm10.8$ & $12.0\pm7.6\hphantom{0}$ & $31.0\pm10.8$ & $44.0\pm8.9\hphantom{0}$ & $28.0\pm7.6\hphantom{0}$ & $36.0\pm2.2\hphantom{0}$ \\
 & Pomaria-1 & $\bestscore{41.0\pm2.2\hphantom{0}}$ & $\hphantom{0}9.0\pm6.5\hphantom{0}$ & $12.0\pm2.7\hphantom{0}$ & $11.0\pm8.2\hphantom{0}$ & $26.0\pm4.2\hphantom{0}$ & $10.0\pm6.1\hphantom{0}$ & $20.0\pm12.2$ \\
 & Pomaria-2 & $\bestscore{57.0\pm5.7\hphantom{0}}$ & $13.0\pm5.7\hphantom{0}$ & $\hphantom{0}6.0\pm4.2\hphantom{0}$ & $\hphantom{0}8.0\pm6.7\hphantom{0}$ & $54.0\pm12.9$ & $18.0\pm7.6\hphantom{0}$ & $17.0\pm10.4$ \\
 & Rs & $41.0\pm5.5\hphantom{0}$ & $\hphantom{0}6.0\pm2.2\hphantom{0}$ & $\hphantom{0}6.0\pm6.5\hphantom{0}$ & $\hphantom{0}8.0\pm6.7\hphantom{0}$ & $26.0\pm7.4\hphantom{0}$ & $\hphantom{0}8.0\pm4.5\hphantom{0}$ & $\bestscore{86.0\pm2.2\hphantom{0}}$ \\
\bottomrule

\end{tabular}
}
\vskip 0.1in
\end{table*}

\subsection{CARLA}
\label{appendix:carla}

For autonomous driving in CARLA, the goal of the agent is to drive as far as possible on an 8-figure highway without collision under diverse weather conditions. We implement the environment in CARLA v0.9.9.4 \cite{dosovitskiy2017carla} and adopt the reward function in \citet{amyzhang2020invariant}:  

\vspace{0.05in}
$ \quad  r_t = \mathbf{v}_\text{agent}^{\top} \hat{\mathbf{u}}_{\text{highway}} \cdot \Delta t - \lambda_c \cdot \text{collision} - \lambda_s \cdot |\text{steer}|  $ 
\vspace{0.05in}

where $\mathbf{v}_\text{agent}$ is the velocity vector of our vehicle, and the dot product with the highway's unit vector $\hat{\mathbf{u}}_{\text{highway}}$ encourages progression along the highway as fast as possible. $\Delta t=0.05$ discretizes the simulation time. We penalize collision, measured as impulse in Newton-seconds, and excessive steering. The respective coefficients are \smash{$\lambda_c=10^{-4}$} and $\lambda_s=1$. We do not investigate more sophisticated rewards like lane-keeping and traffic sign compliance, as they are not the main focus of this paper. We use action repeat 8 for all agents.

\subsection{iGibson}
\label{appendix:igibson}

The goal of the agent in iGibson \cite{xia2020igibson,shen2020igibsonicra} is to find a lamp hanging from the ceiling and navigate to it as closely as possible.  
Our agent is a virtual LoCoBot \cite{gupta2018locobot}. The action dimension is 2, which controls linear velocity and angular velocity. We use continuous action space with 10Hz control frequency. 

Table 6 of the main paper reports the average of 10 \textit{Easy} and 10 \textit{Hard} rooms. We provide a detailed breakdown of those results in Table \ref{table:ig-breakdown} over different floorplans. The \textit{Easy} and \textit{Hard} settings feature distinct interior decorations with different visual distribution shifts from the training room.   

\section{Additional Ablation Studies}

\subsection{\secant-Parallel Results on Robosuite}
\label{appendix:secant-parallel}

We include more experiments with the \secant-Parallel variant on Robosuite (table \ref{table:robo-parallel-variant}) in addition to the DMControl results in Section 5.1 of the main paper. The performance numbers further validate that it is beneficial to train the expert and the student in sequence, rather than in parallel.  

\subsection{More Augmentation Strategies}
\label{appendix:aug-ablations}

In addition to the ablations in Section 5.1, we present extensive results with alternative 2-stage and 1-stage augmentation strategies in Table \ref{table:no-aug-ablation}. The columns ``No-aug $\rightarrow$ Weak" and ``No-aug $\rightarrow$ Strong" are student distillation with weak and strong augmentations, respectively. ``No-aug" column denotes the single-stage policy trained with no augmentation and directly evaluated on unseen tests. ``Strong-only" column is a single-stage policy trained with Combo1 (Section 4.2)  augmentation. \secant outperforms these baselines in a variety of Robosuite and DMControl tasks, which demonstrates that weakly-augmented expert followed by strongly-augmented student is indeed necessary for achieving SOTA performance.

\begin{table}[t!]
\vskip -0.9in
\caption{\secant-Parallel variant on Robosuite. It is advantageous to train expert and student sequentially rather than in parallel.
}
\label{table:robo-parallel-variant}
\vskip 0.1in
\centering
\scriptsize

\resizebox{\columnwidth}{!}{
\begin{tabular}{c|r|cc}
\toprule
Setting & Task & \secant & \secant-Parallel \\ \midrule 

\multirow{4}{*}{\makecell{Robosuite \\ Easy}} & Door opening & $\bestscore{782\pm93\hphantom{0}}$ & $529\pm145$ \\
 & Nut assembly & $\bestscore{419\pm63\hphantom{0}}$ & $374\pm64\hphantom{0}$ \\
 & Two-arm lifting & $\bestscore{610\pm28\hphantom{0}}$ & $390\pm83\hphantom{0}$ \\
 & Peg-in-hole & $\bestscore{837\pm42\hphantom{0}}$ & $540\pm80\hphantom{0}$ \\
 \midrule 
 
 \multirow{4}{*}{\makecell{Robosuite \\ Hard}} & Door opening & $\bestscore{522\pm131}$ & $399\pm71\hphantom{0}$ \\
 & Nut assembly & $\bestscore{437\pm102}$ & $429\pm80\hphantom{0}$ \\
 & Two-arm lifting & $\bestscore{624\pm40\hphantom{0}}$ & $348\pm89\hphantom{0}$ \\
 & Peg-in-hole & $\bestscore{774\pm76\hphantom{0}}$ & $598\pm123$ \\
 \midrule 
 
 \multirow{4}{*}{\makecell{Robosuite \\ Extreme}} & Door opening & $309\pm147$ & $\bestscore{335\pm125}$ \\
 & Nut assembly & $\bestscore{138\pm56\hphantom{0}}$ & $104\pm56\hphantom{0}$ \\
 & Two-arm lifting & $\bestscore{377\pm37\hphantom{0}}$ & $114\pm30\hphantom{0}$ \\
 & Peg-in-hole & $\bestscore{520\pm47\hphantom{0}}$ & $382\pm161$ \\

\bottomrule
\end{tabular}
}

\end{table}
\newcommand{\doublemidrule}[0]{\midrule[0.3pt]\midrule[0.3pt]}

\setlength\doublerulesep{1pt}

\begin{table*}[ht]
\vskip -5in
\caption{Additional ablation studies on alternative augmentation strategies.}
\vskip 0.15in
\label{table:no-aug-ablation}
\resizebox{1.0\textwidth}{!}{%
\begin{tabular}{C{0.1\textwidth}|R{0.15\textwidth}|C{0.15\textwidth}|C{0.15\textwidth}C{0.15\textwidth}C{0.15\textwidth}C{0.15\textwidth}}
\toprule
Setting & Task & \secant & No-aug & No-aug $\rightarrow$ Weak & No-aug $\rightarrow$ Strong & Strong-only \\ \midrule 

\multirow{4}{*}{\makecell{Robosuite \\ \textit{Easy}}} & Door opening & $\bestscore{782\pm93\hphantom{0}}$ & $\hphantom{0}17\pm12\hphantom{0}$ & $\hphantom{0}37\pm21\hphantom{0}$ & $367\pm130$ & $\hphantom{0}47\pm52\hphantom{0}$ \\
 & Nut assembly & $\bestscore{419\pm63\hphantom{0}}$ & $\hphantom{0}3\pm2\hphantom{0}$ & $\hphantom{0}8\pm1\hphantom{0}$ & $172\pm83\hphantom{0}$ & $143\pm95\hphantom{0}$ \\
 & Two-arm lifting & $\bestscore{610\pm28\hphantom{0}}$ & $\hphantom{0}29\pm11\hphantom{0}$ & $\hphantom{0}43\pm15\hphantom{0}$ & $100\pm8\hphantom{00}$ & $\hphantom{0}93\pm26\hphantom{0}$ \\
 & Peg-in-hole & $\bestscore{837\pm42\hphantom{0}}$ & $186\pm62\hphantom{0}$ & $185\pm67\hphantom{0}$ & $489\pm32\hphantom{0}$ & $287\pm63\hphantom{0}$ \\
 \midrule 
 
 \multirow{4}{*}{\makecell{Robosuite \\ \textit{Hard}}} & Door opening & $\bestscore{522\pm131}$ & $\hphantom{0}11\pm10\hphantom{0}$ & $\hphantom{0}31\pm15\hphantom{0}$ & $270\pm94\hphantom{0}$ & $\hphantom{0}36\pm37\hphantom{0}$ \\
 & Nut assembly & $\bestscore{437\pm102}$ & $\hphantom{0}6\pm7\hphantom{0}$ & $13\pm5\hphantom{0}$  & $150\pm40\hphantom{0}$ & $136\pm32\hphantom{0}$ \\
 & Two-arm lifting & $\bestscore{624\pm40\hphantom{0}}$ & $\hphantom{0}28\pm11\hphantom{0}$ & $\hphantom{0}46\pm12\hphantom{0}$ & $99\pm9\hphantom{0}$ & $101\pm37\hphantom{0}$ \\
 & Peg-in-hole & $\bestscore{774\pm76\hphantom{0}}$ & $204\pm81\hphantom{0}$ & $201\pm53\hphantom{0}$ & $353\pm74\hphantom{0}$ & $290\pm81\hphantom{0}$ \\
 \midrule 
 
 \multirow{4}{*}{\makecell{Robosuite \\ \textit{Extreme}}} & Door opening & $\bestscore{309\pm147}$ & $\hphantom{0}11\pm10\hphantom{0}$ & $\hphantom{0}24\pm15\hphantom{0}$ & $190\pm37\hphantom{0}$ & $\hphantom{0}32\pm31\hphantom{0}$ \\
 & Nut assembly & $\bestscore{138\pm56\hphantom{0}}$ & $\hphantom{0}2\pm1\hphantom{0}$ & $5\pm3$ & $\hphantom{0}33\pm11\hphantom{0}$ & $\hphantom{0}43\pm28\hphantom{0}$ \\
 & Two-arm lifting & $\bestscore{377\pm37\hphantom{0}}$ & $25\pm7\hphantom{0}$ & $\hphantom{0}46\pm12\hphantom{0}$ & $\hphantom{0}65\pm13\hphantom{0}$ & $\hphantom{0}58\pm27\hphantom{0}$  \\
 & Peg-in-hole & $\bestscore{520\pm47\hphantom{0}}$ & $164\pm63\hphantom{0}$ & $197\pm66\hphantom{0}$ & $285\pm80\hphantom{0}$ & $289\pm66\hphantom{0}$ \\

\doublemidrule

\multirow{4}{*}{\makecell{DMControl \\ \textit{Color}}} & Cheetah run & $\bestscore{582\pm64\hphantom{0}}$ & $133\pm26\hphantom{0}$ & $\hphantom{0}76\pm23\hphantom{0}$ & $160\pm29\hphantom{0}$ & $296\pm13\hphantom{0}$  \\
 & Ball in cup catch & $\bestscore{958\pm7\hphantom{00}}$ & $151\pm36\hphantom{0}$ & $125\pm26\hphantom{0}$ & $161\pm17\hphantom{0}$ & $777\pm51\hphantom{0}$ \\
 & Cartpole swingup & $\bestscore{866\pm15\hphantom{0}}$ & $248\pm24\hphantom{0}$ & $231\pm31\hphantom{0}$ & $296\pm27\hphantom{0}$ & $628\pm118$ \\
 & Walker walk & $\bestscore{856\pm31\hphantom{0}}$ & $144\pm19\hphantom{0}$ & $\hphantom{0}81\pm13\hphantom{0}$ & $153\pm16\hphantom{0}$ & $598\pm47\hphantom{0}$ \\
\midrule 

\multirow{4}{*}{\makecell{DMControl \\ \textit{Video}}} & Cheetah run & $\bestscore{428\pm70\hphantom{0}}$ & $\hphantom{0}80\pm19\hphantom{0}$ & $\hphantom{0}63\pm15\hphantom{0}$ & $158\pm30\hphantom{0}$ & $271\pm20\hphantom{0}$ \\
 & Ball in cup catch & $\bestscore{903\pm49\hphantom{0}}$ & $172\pm46\hphantom{0}$ & $134\pm33\hphantom{0}$ & $143\pm8\hphantom{0}$ & $727\pm59\hphantom{0}$ \\
 & Cartpole swingup & $\bestscore{752\pm38\hphantom{0}}$ & $204\pm20\hphantom{0}$ & $245\pm17\hphantom{0}$ & $285\pm29\hphantom{0}$ & $503\pm99\hphantom{0}$ \\
 & Walker walk & $\bestscore{842\pm47\hphantom{0}}$ & $104\pm14\hphantom{0}$ & $85\pm11$ & $148\pm15\hphantom{0}$ & $547\pm51\hphantom{0}$ \\
 
\bottomrule
\end{tabular}
}%
\end{table*}

\end{document}